\documentclass[preprint,12pt]{elsarticle}

\usepackage{amsmath,amsfonts}
\usepackage{array}
\usepackage[caption=false,font=normalsize,labelfont=sf,textfont=sf]{subfig}
\usepackage{textcomp}
\usepackage{stfloats}
\usepackage{url}
\usepackage{verbatim}
\usepackage{graphicx}
\usepackage[small]{caption}
\usepackage{arydshln}
\usepackage{amssymb}
\usepackage[inkscapearea=page]{svg}
\usepackage{adjustbox}
\usepackage{multirow}
\usepackage{fancyhdr,graphicx}
\usepackage{hyperref}
\usepackage{fancyhdr,graphicx,amsmath,amssymb}
\usepackage[normalem]{ulem}

\usepackage{float}
\usepackage[ruled,vlined,commentsnumbered]{algorithm2e}
\usepackage[noend]{algpseudocode}

\usepackage{xurl}

\usepackage{pdflscape}
\usepackage{longtable}
\usepackage{afterpage}

\biboptions{sort&compress}

\makeatletter
\def\ps@pprintTitle{%
 \let\@oddhead\@empty
 \let\@evenhead\@empty
 \let\@oddfoot\@empty
 \let\@evenfoot\@empty
}
\makeatother

\begin{document}

\begin{frontmatter}



\title{
Wavelet Probabilistic Recurrent Convolutional Network for Multivariate Time Series Classification
}


\author[1]{Pu Yang}
\author[1]{J.~A.~Barria\corref{cor1}}
\cortext[cor1]{Corresponding author}
\ead{j.barria@imperial.ac.uk}


\affiliation[1]{organization={Department of Electrical and Electronic Engineering, Imperial College London}, 
                addressline={South Kensington Campus}, 
                city={London}, 
                postcode={SW7 2AZ}, 
                country={United Kingdom}}

\fntext[label1]{Tel: +44 (0)20 7594 6275}

\begin{abstract}
This paper presents a Wavelet Probabilistic Recurrent Convolutional Network (WPRCN) for Multivariate Time Series Classification (MTSC), especially effective in handling non-stationary environments, data scarcity and noise perturbations. 
We introduce a versatile wavelet probabilistic module designed to extract and analyse the probabilistic features, which can seamlessly integrate with a variety of neural network architectures.
This probabilistic module comprises an Adaptive Wavelet Probabilistic Feature Generator (AWPG) and a Channel Attention-based Probabilistic Temporal Convolutional Network (APTCN). Such formulation extends the application of wavelet probabilistic neural networks to deep neural networks for MTSC. 
The AWPG constructs an ensemble probabilistic model addressing different data scarcities and non-stationarity; it adaptively selects the optimal ones and generates probabilistic features for APTCN.
The APTCN analyses the correlations of the features and forms a comprehensive feature space with existing MTSC models for classification.
Here, we instantiate the proposed module to work in parallel with a Long Short-Term Memory (LSTM) network and a Causal Fully Convolutional Network (C-FCN), demonstrating its broad applicability in time series analysis.
The WPRCN is evaluated on 30 diverse MTS datasets and outperforms all the benchmark algorithms on average accuracy and rank, exhibiting pronounced strength in handling scarce data and physiological data subject to perturbations and non-stationarities.

\end{abstract}




\end{frontmatter}




\section{Introduction}
Time series (TS) data, one of the most prevalent types of datasets, encompasses crucial physiological signals such as Electrocardiograms (ECGs) and Electroencephalograms (EEGs) that provide insights into cardiac and brain activities. 
A thorough analysis of the trends and patterns within TS data enables the development of resilient forecasting and classification frameworks, which are critical in applications such as financial forecasting, network traffic analysis, and the diagnosis of various physiological conditions \cite{Sezer2020Financial20052019, Barria2011DetectionVariables, Ingolfsson2021ECG-TCN:Network}.

In general, time series classification (TSC) methods can be categorised into two groups, the traditional methods and the Deep Learning (DL)-based methods \cite{IsmailFawaz2019DeepReview}. Traditional methods include techniques such as Dynamic Time Warping with One-Nearest Neighbour (DTW-1NN) and ensemble-based solutions such as decision trees or support vector machines. DL-based methods employ deep neural networks (DNNs) such as Multi-layer Perceptrons (MLPs), Recurrent Neural Networks (RNNs), and Convolutional Neural Networks (CNNs) to discover the complex relationship between the input and its target label. 

DL-based algorithms, which leverage the end-to-end solutions provided by DNNs, reduce the requirement of human interventions and feature engineering, and can generate useful features for specific tasks. The design of the DNNs enables the training on large datasets with various sequence lengths or number of features. Therefore, DNNs have been widely applied in fields such as natural language processing, computer vision, and TSC \cite{Sutskever2014SequenceNetworks, He2016DeepRecognition, IsmailFawaz2019DeepReview}. 

For sequential data, Recurrent Neural Networks (RNNs) are preferable by the nature of their design to capture the temporal dependencies \cite{IsmailFawaz2019DeepReview, Sajjad2020AForecasting}. 
For images or videos, which usually involve high-dimensional data, e.g., images with multiple spatial dimensions or different views of an object for various classification tasks, CNNs are favoured  \cite{Wang2019DevelopmentSurvey}.

Regarding TSC tasks, typical methods involve: (i) using only CNNs for classification \cite{Wang2017TimeBaseline, IsmailFawaz2020InceptionTime:Classification}; or (ii) using CNNs and RNNs in parallel or in sequence to extract temporal features for classification \cite{ Karim2018LSTMClassification, Karim2019MultivariateClassification}.

In particular, Wang \textit{et al.} 
\cite{Wang2017TimeBaseline} showed that the Fully Convolutional Networks (FCNs) outperformed MLPs and Residual Networks in classifying univariate time TS data. Additionally, the research conducted in Fawaz \textit{et al.} \cite{IsmailFawaz2020InceptionTime:Classification} utilised multiple CNNs with residual connections to form an ensemble model for univariate and multivariate TSC tasks. Regarding the utilisation of multiple DNN architectures, the work proposed in Karim \textit{et al.} \cite{Karim2018LSTMClassification} employed LSTM and FCN modules in parallel to extract temporal features for univariate TSC tasks. An advanced version of this approach was later presented in \cite{Karim2019MultivariateClassification}, incorporating the attention mechanism to model the cross-channel relationships of the MTS data to improve the model performance for MTSC tasks.

Wavelet signal processing, a powerful signal processing technique, can generate interpretable representations for TS data analysis, as it has good localisation capability in the time and frequency domains, and can decompose a signal into different resolutions via wavelet transforms \cite{Chaovalit2011DiscreteMining}. 

Most wavelet-based approaches use wavelet transform (WT) to denoise data or decompose data into different scales and explore new feature representations for TS data analysis (see for example, \cite{Zhang2004AClassification, Chaovalit2011DiscreteMining, Wang2018MultilevelAnalysis}). 
Regarding DL-based approaches, Wang \textit{et al.} \cite{Wang2018MultilevelAnalysis} decomposed TS data into high- and low-frequency sub-series, containing different information. A residual classification flow and a Long Short-Term Memory (LSTM) network were employed for the classification and prediction tasks. Furthermore, in Pancholi \textit{et al.} \cite{Pancholi2023SourceSignal}, EEGs were decomposed into sub-series using wavelet packet decomposition; the extracted features were employed by the CNNs and LSTMs for classification. Moreover, in Li \textit{et al.} \cite{Li2020WaveletClassification}, the pooling layers in CNN-based models were replaced by wavelet transforms to extract features for noise-robust image classification tasks.

We note that apart from utilising wavelets in the time-frequency domain for feature extraction and denoising, 
wavelets can provide a new perspective on the probability domain through Wavelet Density Estimators (WDEs). A WDE estimates the probability density function of the data, which encapsulates the statistical information of the data, and can be further used for TS data analysis \cite{Garcia-Trevino2014StructuralClassification}. 

WDEs exhibit good localisation with the time and frequency domains, maintain robustness against noise, and outperform alternative methods when addressing densities with local discontinuities. However, a critical factor in the utilisation of WDEs is their computational complexity; although various WDEs are available for modelling data distributions, we should be aware that not all WDEs are endowed with analytical solutions and low computational complexity, which often necessitates considerable computational resources. This limitation also hinders further research on WDEs, particularly concerning Multivariate Time Series (MTS) data, where the computational complexity increases significantly as the dimensionality of features increases.

In Garcia-Trevino \textit{et al.} \cite{Garcia-Trevino2024WaveletNetworks}, we developed the first generative wavelet-based probabilistic neural network (WPNN) for data stream classification and anomaly detection tasks. The key aspect of WPNN is a novel WDE that has analytic solutions with constant computational complexity, and is capable of handling data streams with unconstrained length in both stationary and non-stationary environments. 

In this paper, we propose a feature fusion-based framework, namely Wavelet Probabilistic Recurrent Convolutional Network (WPRCN), for multivariate time series classification (MTSC). WPRCN is particularly suited for 
classifying MTS data subject to noise perturbations, non-stationarity, and data scarcity. Note that non-stationarity refers to the statistical properties of the data changing over time; therefore, traditional classification models fail to capture the change and the classification performance degrades. Such challenges can easily be found in physiological data such as EEGs and ECGs \cite{AlZoubi2015AffectClassifiers}.

We further note that for MTS data, the cross-channel relationships among features at the same timestamp are often enhanced through the channel attention (CA) mechanism (see, for example, \cite{Karim2019MultivariateClassification, Chen2022DA-Net:Classification}). Therefore, the proposed framework also integrates the CA mechanism for feature enhancement.

The proposed WPRCN is graphically depicted in Fig. \ref{fig:system_diag}, detailed in Section \ref{sec:method}. 
It considers three different feature extraction modules; the most important one is the proposed probabilistic module, which consists of an Adaptive Wavelet Probabilistic Feature Generator (AWPG) module and a Channel Attention-based Probabilistic Temporal Convolutional Network (APTCN)-based probabilistic feature analyser module. We further instantiate two DL-based modules to demonstrate the seamless integration capability of the probabilistic module. (ii) an LSTM module, and (iii) a Causal\footnote{Note that the term \textit{Causal} means that the CNN kernels do not have access to any future information.}-FCN (C-FCN) module.

By aggregating features extracted from these three modules, a comprehensive feature representation for MTSC is created for classification.

The contributions of the work are threefold: (i) We propose a novel probabilistic module, comprising the Adaptive Wavelet Probabilistic Feature Generator (AWPG) and the Channel Attention-based Probabilistic Temporal Convolutional Network (APTCN). The AWPG module 
generates an ensemble model characterising different rates of non-stationarity. 
It adaptively selects the optimal probabilistic model from all the possible candidate models, and automatically generates the corresponding features considering different amounts of data availability and different rates of data variation in a non-stationary environment for further analysis. 
(ii) The AWPG addresses the \textit{curse of dimensionality}, a prevalent challenge in prior WDE-based approaches for MTSC, by constructing and modelling a latent space that not only simplifies the complexity of the data but also characterises the key features and dynamics of the original MTS data. In this sense, it differs from prior WDE-based approaches, as it guarantees a bounded and low computational complexity.
(iii) The generated probabilistic features with different smoothness characterising different levels of data scarcity are then analysed by the APTCN, which explores the cross-channel relationships among these features and generates enhanced features for classification purposes.

In this paper, the proposed probabilistic module has been seamlessly integrated with the LSTM and C-FCN networks for the MTSC task. It is important to highlight that this probabilistic module can seamlessly be incorporated into a broad range of DNN architectures, and as a consequence, further enhance the analysis of temporal data.

To the best of our knowledge, WPRCN is the first wavelet probabilistic deep neural network that considers the wavelet probabilistic module for MTSC tasks, as conventional WDEs are limited by (i) lacking analytic solutions for WDEs, (ii) inability to handle data with high dimensionality, (iii) lacking solutions to handle the cross-channel relationships in an MTS data.

\section{Related research}\label{sec:background}

\subsection{Deep Neural Networks for MTSC}

Time series classification (TSC) tasks with neural networks often involve capturing different aspects of features from the time domain \cite{IsmailFawaz2019DeepReview}, in which Multi-layer Perceptrons (MLPs), CNNs, and RNNs are commonly used. 

For univariate time series data, in Wang \textit{et al.} \cite{Wang2017TimeBaseline}, three different network architectures, i.e., the Multi-Layer Perceptron (MLP), the Fully Convolutional Network (FCN) and the Residual Network (ResNet) are used for classification. 

A similar approach, LSTM-FCN, proposed in \cite{Karim2018LSTMClassification}, uses CNNs and RNNs in parallel to extract temporal features from two radically different networks. Moreover, the approach in \cite{HajizadehTahan2022DevelopmentClassification} extracts discretised temporal features using CNNs and combines them with other DNNs for classification.

For MTS data, approaches mainly focus on feature fusion-based solutions, which aim to explore the local and global feature dependencies across the time axis and feature channels using different network architectures. In \cite{Karim2019MultivariateClassification}, a multivariate LSTM-FCN (MLSTM-FCN) utilises the Squeeze-and-Excitation (SE) module to enhance specific feature dimensions from the MTS data, as they contain more informative features for classification. Similarly, the SE module and a self-attention module are used in \cite{Chen2022DA-Net:Classification} to capture the local and global dependencies. In \cite{He2023Rel-CNN:Classification}, a CNN-based module is used to learn the global and local temporal features from the subsequences of the original data. A multi-scale attention CNN proposed \cite{Chen2021}, captures temporal features at different scales for MTSC. In \cite{Yue2025}, a graph attention mechanism with CNNs is used to model the MTS for classification.

Furthermore, a low-dimensional feature space was learnt via LSTMs and CNNs in \cite{Zhang2020TapNet:Network} for classification. In the work proposed in \cite{Tang2022Omni-ScaleClassification}, an Omni-scale block was developed for the 1D-CNN to change the kernel size based on the data, tailored for MTSC tasks. Moreover, in \cite{Tahan2023AClassification}, CNNs were used to extract temporal features, which were then used by other publicly available classifiers, such as the MLSTM-FCN developed in \cite{Karim2019MultivariateClassification}, for classification. Additionally, the work proposed in \cite{Wu2022TimesNet:Analysis} extracted features based on the frequency components and amplitudes of the MTS data using the Fourier Transform, and further utilised CNNs for MTSC. Apart from utilising DNNs, \cite{Baldan2021MultivariateRepresentation} generated a set of informative features, such as skewness and linearity, that can improve the performance of conventional classifiers such as the support vector machine and random forest. 

\subsection{Temporal Convolutional Networks}

In recent years of temporal data modelling, CNN-based architectures, specifically fully convolutional networks (FCNs), have attracted more attention compared to recurrent neural networks (RNNs) that may suffer from vanishing/exploding gradient problems.

The \textit{WaveNet} proposed in \cite{Oord2016WaveNet:Audio}, models the temporal correlation using CNNs and generates possible audio output for the next timestamp. The design of \textit{WaveNet} is complicated and involves skip-connection and gated activations. In \cite{Bai2018AnModeling}, a simpler architecture, Temporal Convolutional Network (TCN), is proposed, in which the FCN layers with causal padding are used. By having dilated causal convolutions with residual connections, the proposed TCN can model longer sequences. 
As regards feature fusion-based MTSC approaches discussed in \cite{Karim2019MultivariateClassification, Chen2022DA-Net:Classification, Zhang2020TapNet:Network, Tang2022Omni-ScaleClassification, Baldan2021MultivariateRepresentation}, few of them consider the information leakage in the FCN module.

As discussed in \cite{Bai2018AnModeling}, TCN and its related CNN-based architectures, can (i) enable parallelism for model training and inference; (ii) provide flexible receptive fields by changing the kernel size and depth of the network; (iii) have more stable gradients than RNNs; and (iv) deal with longer sequences. Hence, their suitability for modelling temporal data renders TCN an appropriate component as a feature analyser module in the proposed WPRCN.

\subsection{Channel Attention}

Attention mechanisms in neural networks compute weights for different parts of a sequence based on their contribution to the tasks. By focusing on important parts and ignoring irrelevant ones, the enhanced features improve the performance in various tasks, including machine translation, image classification, and speech recognition \cite{Brauwers2021ALearning}. There are several types of attention mechanisms, such as self-attention, spatial attention, and channel attention, which are used in different tasks including the sequence and image modelling \cite{Vaswani2017AttentionNeed, Woo2018CBAM:Module, Wang2020ECA-Net:Networks}.

For MTSC tasks, Channel Attention (CA) can be used to explore the inter-channel relationship among the features by selectively adjusting the channel importance based on the contribution to the labels, and therefore, can be treated as a feature enhancement technique. There exist several types of CA, for example, Squeeze-and-Excitation and Efficient Channel Attention \cite{Hu2020Squeeze-and-ExcitationNetworks, Wang2020ECA-Net:Networks}.

The Squeeze-and-Excitation (SE) module shows strong performance in MTSC tasks (see, for example, \cite{Karim2019MultivariateClassification, Chen2022DA-Net:Classification}) by first generating the channel-wise statistics through global average pooling, then re-calibrating the statistics to capture the channel-wise dependencies. 

The Efficient Channel Attention (ECA) module is a lightweight CA approach that, compared to the SE module, requires fewer network parameters and computations; and the cross-channel interaction can be studied using the CNNs without additional dimensionality reduction steps \cite{Wang2020ECA-Net:Networks}. Owing to its efficiency, the ECA module is integrated into the proposed WPRCN for feature enhancement in the probabilistic module.

\subsection{Wavelet Time Series Description}
Wavelets are widely used in signal processing and TSC tasks for denoising, compression, and time-frequency domain analysis \cite{Mallat2008AWay}. For TS data, wavelets can provide two perspectives, i.e., the time-frequency domain and the probability domain. The former can be formed by Wavelet Transform (WT) which decomposes data into different resolutions. WT has been widely applied in medical, remote sensing, or finance fields \cite{Rhif2019WaveletReview}. The latter perspective comes from Wavelet Density Estimators (WDEs), in which the TS data is treated as a stochastic process such that it can be characterised by its probability density function. 

WDE is a non-parametric density estimator with good localisation in time and frequency, and it has superior density estimation performance against other density estimators. Given a group of random variables $X_i$, $i \in [1,N]$, with an unknown square-integrable density function $p(x)$; $p(x)$ can be approximated by WDEs and given as $\quad \hat p(x)= \sum_k \hat a_{j_0,k} \phi_{j_0,k}(x)$, using a scaling function $\phi$. Note that $\hat a_{j_0,k}=  \frac{1}{N} \sum_{i=0}^{N} \phi_{j_0,k}(X_i) $. Despite its outstanding performance in density estimation, conventional WDE has a noticeable limitation that restricts thorough research with DNNs, i.e., the high computational cost due to no analytic forms available. 

The WDE we proposed in \cite{GARCIATREVINO2019111} has closed-form solutions with constant time complexity, in which only the latest data point is stored and used to update the estimator. For TS data in a non-stationary environment, the statistical properties might evolve and hence the existing models require adaptation to such shifts. This approach can adapt to the different rates of statistical properties change in a non-stationary environment, and hence, is selected for the probabilistic feature generation process.

\section{Wavelet Probabilistic Neural Networks}\label{sec:wpnn}

\subsection{Wavelets in Probabilistic Neural Networks}
Wavelet transform (WT) is one of the typical feature extraction components in DNNs. In \cite{Yu2018OnlineNetworks}, WT is used together with a GRU to extract the time-frequency features for fault detection in a smart grid. In \cite{Kanarachos2017DetectingTransform}, the signal is decomposed into different resolutions using WT to denoise the data and the anomaly detection is then done by using DNNs.

Our work in \cite{Garcia-Trevino2024WaveletNetworks}, utilising a WDE with constant time complexity and analytic solutions, is designed for data stream classification in a stationary and non-stationary environment, and shows superior performance compared to existing density-based classifiers.

We note that while any WDEs could potentially be integrated into the proposed probabilistic module in WPRCN, our selection criteria are governed by the computational cost, the availability of an analytic solution, and the capability to model non-stationary environments. In this sense, we have selected the WDE solution proposed in \cite{GARCIATREVINO2019111}, which will be briefly reviewed in the following sections.

\subsection{Radial $B$-spline Scaling Function}

The Radial $B$-spline scaling function $\Phi$ is the fundamental block of the wavelet probabilistic networks, which has analytic solutions. 
Given an input $\mathbf{x}$, which is normalised in the interval of $[0,1]^n$, for any dilation and translation parameter $j_0$ and $k$, the function $\Phi$ is given as:

\begin{equation} 
\Phi_{j_0,k}(\mathbf{x})=2^{\frac{nj_0}{2}}N_m \big( \lVert (2^{j_{0}}\mathbf{x}-\mathbf{k}) \rVert + \frac{m}{2}  \big)
\label{eq:bspline}\end{equation}

\noindent
where $n, j_0, \mathbf{k} \in \mathbb{Z}$ are the dimension of the input $\mathbf{x}$, dilation parameter and the vectorised translation parameter, respectively. The closed-form solution of $N_m(x)$ with different order $m$ is defined as $\phi(x)$ and shown in Table \ref{tbl:closed_form}.

The translation parameters $k$ are defined as $k \in \{-u, ..., 0, 1, 2, ..., 2^{j_0}+u\}$, where $u=1$ for \textit{Linear} and \textit{Quadratic} order and $u=2$ for \textit{Cubic} order. Therefore, the multidimensional version of the translation parameter, $\mathbf{k} \in \mathbb{R}^{l,n}$, depends on data dimension $n$, and can be constructed by creating all the $l$ combinations of $k$ across the feature dimension $n$.

\begin{table}[!ht]
\centering
\caption{Closed-form solution for $B$-spline functions.}
\label{tbl:closed_form}
\scalebox{0.9}
{
\begin{tabular}{lll}
\hline
\textbf{Order}& \textbf{$B$-spline function} \\
\hline
\textit{Linear, $m=2$}

& 
\(
\phi(x)=\begin{cases}x & \hphantom{1111111111111111111}\text{for }0\leq x<1 \\
2-x & \hphantom{1111111111111111111}\text{for }1\leq x<2 \\
0 & \hphantom{1111111111111111111}\text{otherwise } \\
\end{cases}\nonumber
\)
\\
\hline
\textit{Quadratic, $m=3$}
& 
\(
\phi(x)=\begin{cases}\frac{1}{2}x^2 & \hphantom{111111111111}\text{for }0\leq x<1 \\
\frac{3}{4}-(x-\frac{3}{2})^2 & \hphantom{111111111111}\text{for }1\leq x<2 \\
\frac{1}{2}(x-3)^2 & \hphantom{111111111111}\text{for }2\leq x<3 \\
0 & \hphantom{111111111111}\text{otherwise } \\
\end{cases}\nonumber
\)
\\
\hline
\textit{Cubic, $m=4$}
& 
\(
\phi(x)=\begin{cases}
\frac{1}{6}x^{3} & \text{for }0\leq x<1 \\
\frac{1}{6}(-3x^{3}+12x^2-12x+4) & \text{for }1\leq x<2 \\
\frac{1}{6}(3x^{3}-24x^2+60x-44) & \text{for }2\leq x<3 \\
\frac{1}{6}(4-x)^{3} & \text{for }3\leq x<4 \\
0 & \text{otherwise } \\
\end{cases}\nonumber
\)
\\
\hline
\end{tabular}
}
\end{table}

\subsection{Density Estimation}\label{sec:density_esti}
An unknown square integrable density function $p(\mathbf{x})$ can be represented as $\hat p(\mathbf{x}) = \sum_{k} \hat w_{j_{0},k} \Phi_{j_{0},k}(\mathbf{x})$. Given a group of random variables $\{X_i\}_{i \in [1, N] }$, the estimated network coefficient $\hat{w}$ for any $j_0, k$ can be computed as $\hat w=\frac{1}{N} \sum_{i=0}^{N} \Phi(X_i)$, and can be further optimised into the following form:

\begin{equation} 
\begin{aligned}
\hat w &=(1-\alpha_\Gamma)\hat w + \alpha_\Gamma\Big( 2^{\frac{nj_0}{2}}\phi\big( \lVert (2^{j_{0}}\mathbf{x}-\mathbf{k}) \rVert + \frac{m}{2} \big)  \Big)
\end{aligned}
\label{eq:online_nonstat}
\end{equation}

\noindent
where $\alpha_\Gamma = \{\alpha_1,\alpha_2,...,\alpha_\gamma\}$ 
is a set containing different scales of the receptive field $\alpha$; $\alpha$ refers to the size of the receptive field, which can also be interpreted as a forgetting factor that is inversely related to the size of a sliding window; note that we set $\alpha_\Gamma = \{1, 1/10, 1/100, 1/500, 1/1000\}$ in the paper; $\hat{w}\in \mathbb{R}^{|k|, |\Gamma|}$ is the network parameters, where $|\cdot|$ is the cardinality.

A higher value of $\alpha$ is expected when the data's statistical properties change rapidly, which corresponds to a smaller window that highlights more recent data. Hence, computing $|\Gamma|$ views that describe different rates of data variation can be perceived as having multiple receptive fields, and an ensemble probabilistic model of $|\Gamma|$ different perspectives can be formed for further analysis.

Equation (\ref{eq:class_prob}) computes the probability density ensemble of the input after the network coefficient $\hat{w}$ is trained: 

\begin{equation} 
\begin{aligned}
\hat p(\mathbf{x})
&= \sum_{l}{\hat{w}_l \cdot 2^{\frac{nj_0}{2}}}  \phi(\left\Vert 2^{j_0}\mathbf{x}-\mathbf{k}_l \right\Vert+\frac{m}{2})
\end{aligned}
\label{eq:class_prob}
\end{equation}

\noindent
where $\hat p(\mathbf{x})$ contains $|\Gamma|$ different perspectives.

\section{Methodology}\label{sec:method}

\begin{figure*}[!htp]
\centering
\includegraphics[trim = 0mm 0mm 0mm 0mm,clip,width=0.8\linewidth]{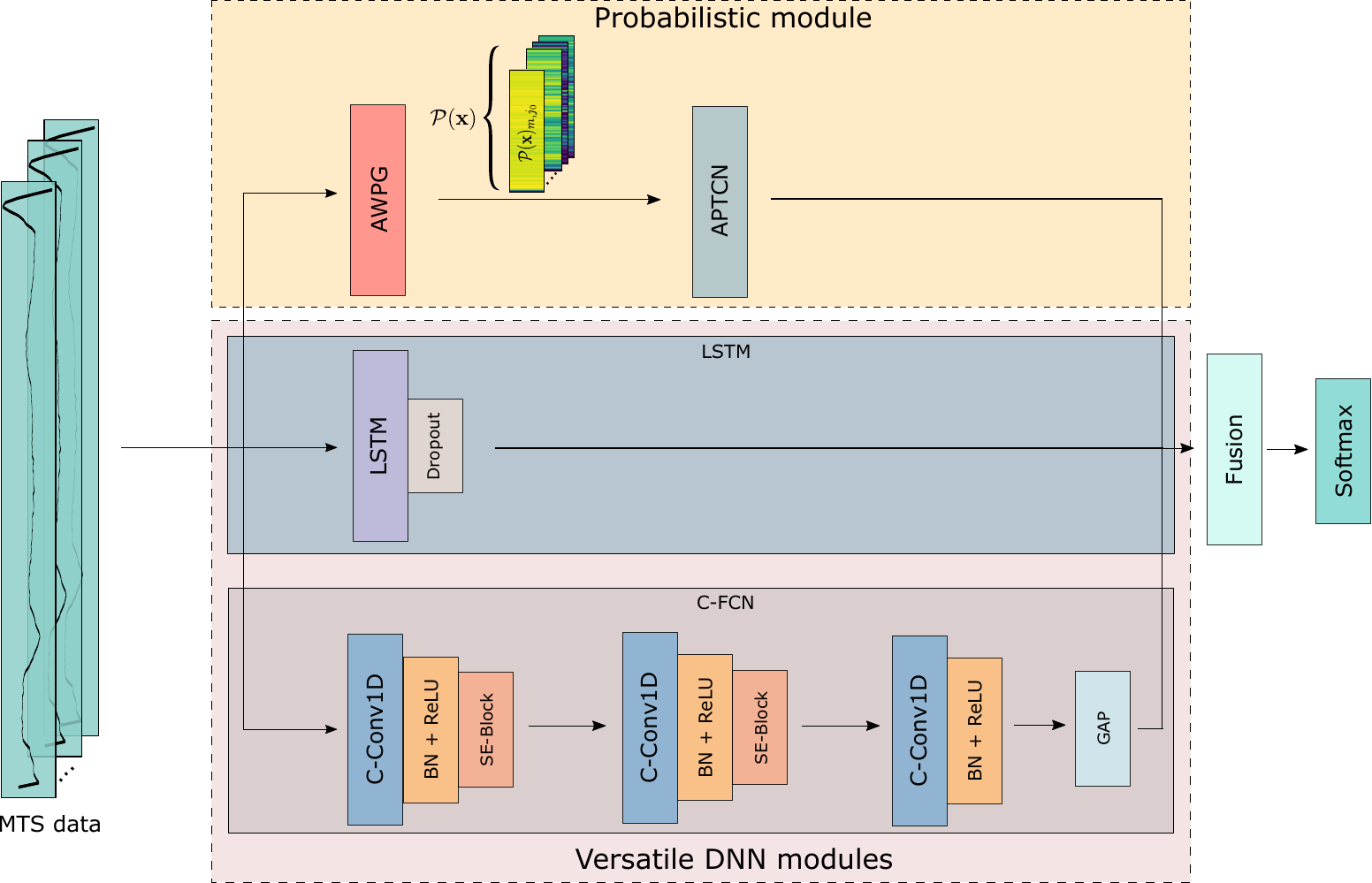}
\centering
\caption{
The proposed WPRCN workflow.
The novel Probabilistic module can work with a variety of neural network architectures. An LSTM and a C-FCN are employed to show the broad applicability of the probabilistic module for MTSC TASKS.
For the probabilistic module (AWPG and APTCN), the data is transformed into a probabilistic feature 
ensemble
$\mathcal{P}(\mathbf{x})$ with different smoothness considering different levels of data scarcity and non-stationarity using AWPG. The APTCN receives $\mathcal{P}(\mathbf{x})$ and utilises the channel attention to learn the enhanced feature representations. Such representations are then subjected to further analysis within the APTCN pipeline for classification.
The features collated from each feature extraction module are consolidated at the Fusion layer, and passed to a Softmax layer for classification.
}\label{fig:system_diag}
\end{figure*}

The proposed WPRCN contains three distinctly different feature extraction modules, which are shown in Fig. \ref{fig:system_diag}. In the figure, the three parallel modules, i.e., (i) the novel probabilistic module, which contains a feature generator (AWPG) and an analyser (APTCN); (ii) an LSTM; and (iii) a 1D-Causal FCN (C-FCN) network module with SE module and Global Average Pooling (GAP) layer for temporal feature extraction, are shown in sequence from top to bottom. The features from these three modules are combined and transformed in a \textit{Fusion} layer and passed to a Softmax layer for classification.

We would like to highlight that the probabilistic module can be seamlessly integrated with a variety of neural network architectures for temporal data analysis. In this paper, we employed an LSTM and a C-FCN to instantiate its broad applicability for MTSC.

In this section, we will first introduce the components within the proposed probabilistic module, detailing the Adaptive Wavelet Probabilistic Feature Generator (AWPG) and the Channel Attention-based Probabilistic Temporal Convolutional Network (APTCN). Subsequently, an overview of all other associated modules of the WPRCN will be provided.

\subsection{Adaptive Wavelet Probabilistic Feature Generator (AWPG)}

The AWPG generates the probabilistic features and consists of three components: (i) A GRU-based Encoder-Decoder (GED) module, which learns a latent space for the input MTS data, (ii) an ensemble Multi-Receptive-field Wavelet Probabilistic Network (MRWPN), which creates an ensemble model for the latent space. This is achieved by estimating the probability density functions through a novel wavelet density estimator in a non-parametric manner, notable for its low computational complexity and adaptability to non-stationary environments, and (iii) an \textit{Adaptive Network}, which predicts the optimal component $I$ that is associated with a specific rate of data variation from the ensemble views.

Fig. \ref{fig:ARWPNN} shows a detailed illustration of the proposed AWPG: the input MTS data $\mathbf{x}$ is fed into the Encoder of GED, which forms a latent space characterised by features $h^E_L$ and $\text{\textbf{y}}_E$. Subsequently, the Decoder of GED utilises $h^E_L$ for data reconstruction. In parallel, $\text{\textbf{y}}_E$ is passed to MRWPN (parameterised by specific $m$ and $j_0$) for generating the feature $\hat{p}(\text{\textbf{y}}_E)$. The predicted index, $I$, is then employed to select the optimal feature, which is denoted as $\mathcal{P}(\mathbf{x})_{m,j_0} = \hat{p}_I(\text{\textbf{y}}_E)$.

\subsubsection{GRU-based Encoder Decoder (GED)}

Consider an MTS data $\mathbf{x} = [x_1, x_2, …, x_L]$, $x_i = [x^1_i, x^2_i, …, x^n_i]$, where $L$ is the length of the data, $n$ is the feature dimension of the data in each timestamp. The latent features, i.e., $h^E_L$ and $\text{\textbf{y}}_E$, are generated by the GED module, which consists of $E$ GRU layers in the Encoder, and $D$ GRU layers in the Decoder, $ [E, D]\in\mathbb{Z}$. 
The output $\text{\textbf{y}}_{e-1}$ of the GRU in the Encoder at layer $e-1$ is passed to the next GRU,  $e = [1,2,..., E]$, and the final hidden state $h^E_L$  from the last GRU layer in the Encoder is passed to Decoder for data reconstruction. Meanwhile, the output $\text{\textbf{y}}_E$ and the final hidden state $h^E_L$ are fed into MRWPN for probabilistic feature generation and latent space modelling tasks, respectively. Note that we define \textit{latent space modelling} as the process of building a one-class classifier utilising $h^E_L$, which encapsulates the entire historical context of the input.

\subsubsection{MRWPN}

The probabilistic features, denoted as $\hat{p}(\text{\textbf{y}}_E)$, are generated by MRWPN, employing specific values of $m$ and $j_0$. These two parameters ($m$ and $j_0$) control the degree of smoothness of the estimated densities and accommodate varying amounts of training data. Note that in scenarios with limited training data, lower values of $m$ and/or $j_0$ are preferred. Furthermore, MRWPN generates $|\Gamma|$ distinct models, each encapsulating a unique rate of data variation (for an in-depth formulation of MRWPN, refer to Section \ref{sec:density_esti}).

\begin{figure}[htp]
\centering
\includegraphics[trim = 0mm 0mm 0mm 0mm,clip,width=\linewidth]{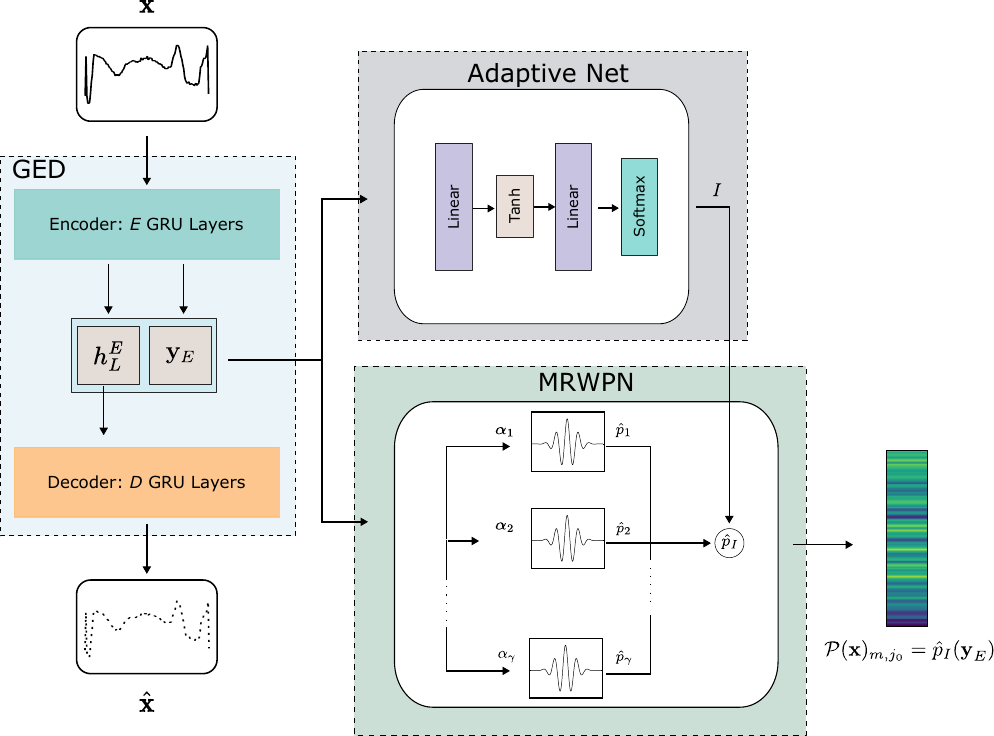}
\caption{The proposed Adaptive Wavelet Probabilistic Feature Generator (AWPG), which contains (i) a GRU-based Encoder-Decoder to form a latent space, (ii) a MRWPN that models the latent space and provides multiple views against different rates of data variation, and (iii) an \textit{Adaptive Network} that predicts the optimal index $I$ for the selection of a probabilistic model from the ensemble views.
}\label{fig:ARWPNN}
\end{figure}

\subsubsection{Adaptive Network}

The key to the proposed AWPG is the adaptive model selection capability that allows for the automatic determination of the optimal model from a set of candidate models, which is crucial in generating features suited for classification. Given a set of indices for the candidate models, $\Gamma = \{1,2,...,\gamma\}$ 
, the \textit{Adaptive Network} predicts the optimal index $I$ that is associated with a higher value of $\hat{p}_I(h^E_L)$.

The AWPG is trained in an unsupervised way, modelling one class of the data. 
Therefore, we expect more distinctions in $\mathcal{P}(\mathbf{x})_{m,j_0}$ between the remaining classes and this class (used for model training).
One key aspect of the MRWPN module is that the network weights $\hat{w}$ can be updated in one pass without back-propagation $\left( \text{see Equation (\ref{eq:online_nonstat})} \right)$.
In order to associate the MRWPN and \textit{Adaptive Network} with the network training process, the term $\hat{p}(h^E_L)$ is added to the loss function:

\begin{equation}
\mathcal{L} = \frac{1}{B_s} \sum_{i=1}^{B_s} |\mathbf{x}_i - \hat{\mathbf{x}}_i| - \frac{\lambda}{B_s} \sum_{i=1}^{B_s} \log(\hat{p}({h_i}^E_{L})*\theta^I_i)
\label{eq:loss_func}
\end{equation}

\noindent
where $B_s$ is the batch size, $\lambda$ is a tunable parameter that controls the focus of the loss function, and $\theta^I$ are the predicted weights from the \textit{Softmax} layer in the \textit{Adaptive Network}. Note that this joint optimisation simultaneously addresses the construction and modelling tasks of the latent space.

Prior to the generation of the probabilistic features, it is essential to find the best model configurations of AWPG. The model performance, with a specific focus on latent space modelling, is evaluated in a one-class classification task utilising $h^E_L$, and can be measured by a series of classification metrics. 
Hence, the optimal configurations of the GED, Adaptive Network, MRWPN, and a threshold $\beta$ can be found by optimising an equation such that:

\begin{equation}
\operatorname*{argmax}_{\beta} \text{F1}\{\hat{p}_I(h^E_L) < \beta\}
\label{eq:decision_alpha_update}
\end{equation}

\noindent
where $I$ is the predicted index generated by the \textit{Adaptive Network}; F1 is the F1-score utilised for evaluating the identification of the unseen class of data; $\beta$ is a threshold. Note that the applicability of this equation extends beyond its present utilisation, encompassing anomaly detection tasks, which we have empirically examined. However, these findings fall outside the scope of this paper and, hence, will not be covered herein. These additional insights will be reported in a subsequent publication.

Regarding the generation of the probabilistic feature, the Encoder output $\text{\textbf{y}}_E = Enc(\mathbf{x})$ from the pre-trained GED along with the associated optimal index $I$ are used to train the MRWPN module employing a new set of the network weight $\hat{w}$ $\left( \text{see Equation (\ref{eq:online_nonstat}}) \right)$. The resulting probabilistic feature $\mathcal{P}(\mathbf{x})_{m,j_0}$ with the $m$-order and the resolution $j_0$ for the input $\mathbf{x}$ can be generated using Equation (\ref{eq:class_prob}) such that $\mathcal{P}(\mathbf{x})_{m,j_0} = \hat{p}_{I}(\text{\textbf{y}}_E)$.

Each feature, $\mathcal{P}(\mathbf{x})_{m,j_0}$, is associated with two hyperparameters, i.e., the order of $B$-spline $m$, and the resolution parameter $j_0$. These two parameters control the smoothness of the estimated densities and depend on the data availability. Hence, varying the values of $m$ and $j_0$ can uncover different levels of information from the probability domain.
Thus, we include features with all the combinations of $m$ and $j_0$, and create $\mathcal{P}(\mathbf{x})$ with $\mathcal{C} = |M|\cdot|J|$ channels, where $M$ is the pre-defined set of three different orders of the $B$-spline, $J = \{1,2,3,4,5\}$ is the set of resolution parameters.

Given our primary focus on the entire probabilistic module, for simplicity, we fix the Encoder output of GED to $2$, and a two-layer architecture of the encoder and decoder output feature sizes are searched from the space $\{128,64,32,16,8,4\}$ in ascending and descending order, respectively. We set $\lambda$ to $0.1$ to balance the two terms in Equation (\ref{eq:loss_func}). 
Regarding the network architecture of the \textit{Adaptive Network} (see Fig. \ref{fig:ARWPNN}), a \textit{Linear} layer with an output size of $10$ is used; this is followed by a \textit{Tanh} activation function and a subsequent \textit{Linear} layer with an output size of $\Gamma$; and finally, a \textit{Softmax} layer is used to predict the associated index.

\begin{figure}[htp]
\centering
\includegraphics[trim = 0mm 0mm 0mm 0mm,clip,width=0.9\linewidth]{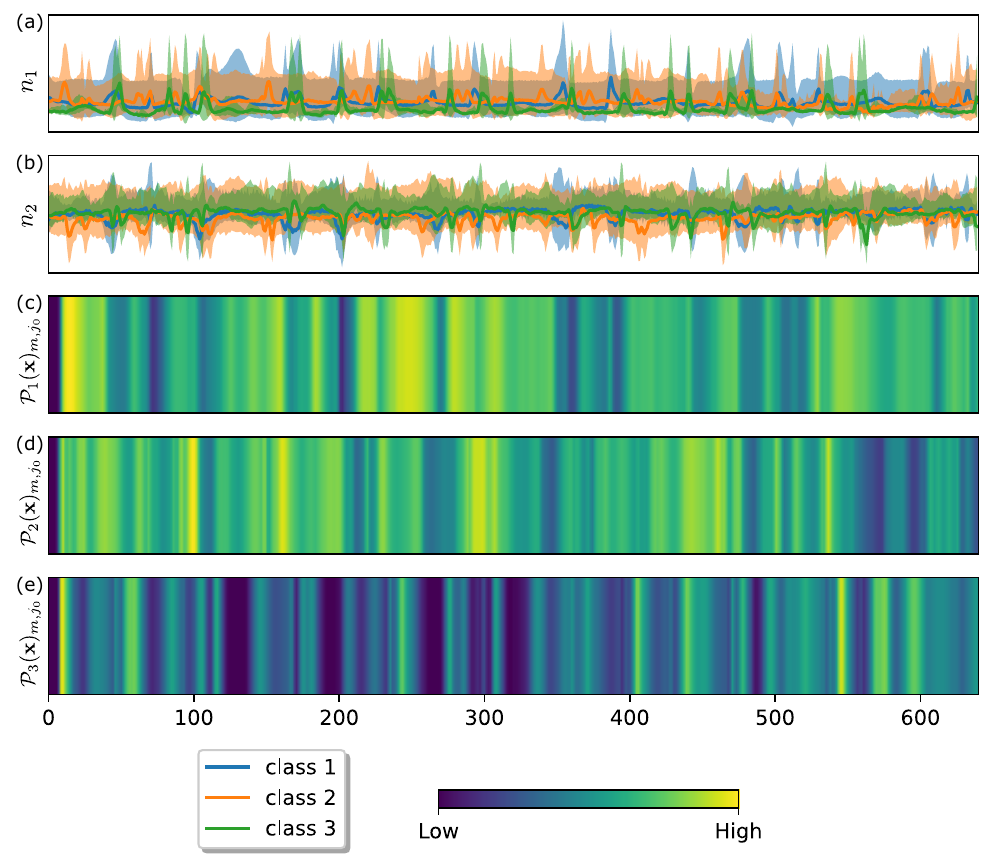}
\caption{Visualisation of features in the \textit{AtrialFibrillation} dataset. (a) and (b) represent the two feature dimensions of the data, illustrating the class-wise means with shaded regions indicating the respective deviations. (c-e) show the three distinct probabilistic features denoted as $\mathcal{P}_c(\mathbf{x})_{m,j_0}$ for each of the classes using $m$ and $j_0$.} \label{fig:pdf_demo}
\end{figure}

In Fig. \ref{fig:pdf_demo}, we show an example of a two-dimensional source signal and its corresponding probabilistic features $\mathcal{P}(\mathbf{x})_{m,j_0}$ for dataset \textit{AtrialFibrillation} from the repository UEA proposed in \cite{Bagnall2018The2018}. 
Note that we train the AWPG using the first class ($y=1$) in the dataset, therefore, $\mathcal{P}(\mathbf{x})_{m,j_0}$ refers to the probabilistic pattern for data with $y=1$.
It is a two-channel ($n=2$) ECG recording collection of three different atrial fibrillations with $L=640$. 

Fig. \ref{fig:pdf_demo}(a-b) presents the corresponding feature dimensions $n_1$ and $n_2$, respectively. The shaded regions surrounding the mean signal of each class are the standard deviations, effectively encapsulating the variability inherent to each atrial fibrillation pattern. 
Fig. \ref{fig:pdf_demo}(c-e) are the associated probabilistic features for each of the three classes. These three features exhibit notable variations in values and smoothness, and therefore, they encapsulate key insights for delineating class boundaries for MTSC tasks. 

We further note that the usage of the wavelet probabilistic module in MTSC tasks has not been researched in the context of DNNs due to the absence of suitable WDE solutions.

\subsection{Channel Attention-based Probabilistic Temporal Convolutional Network (APTCN)}

Following the generation of the probabilistic features, the proposed APTCN, undertakes a thorough analysis of these features for classification. 

\begin{figure*}[!htp]
\centering
\includegraphics[trim = 0mm 0mm 0mm 0mm,clip,width=0.9\linewidth]{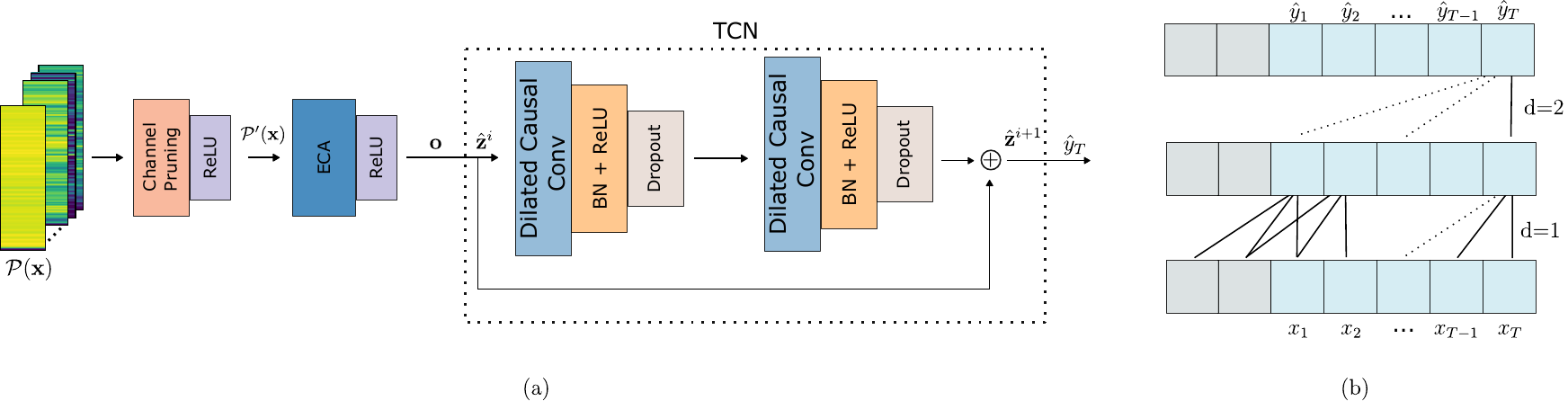}
\centering
\caption{
The workflow of the proposed APTCN. (a) A dilated causal convolutional network for modelling $\mathcal{P}(\mathbf{x})$, which involves a channel pruning stage and a channel attention stage; the output $\mathbf{o}$ is then passed to the TCN. For the TCN, the input at depth $i$ is given as $\hat{\mathbf{z}}^i$ and the output $\hat{\mathbf{z}}^{i+1}$ is used for the next level. The output of the TCN, denoted as $\hat{y}_T$, is used for classification. (b) The detailed visualisation of a $2$-level Dilated Causal Convolution with $\mathbb{K}=3$ and $d=1, 2$, where the grey blocks are the zero-paddings.
}\label{fig:aptcn_diag}
\end{figure*}

Fig. \ref{fig:aptcn_diag} illustrates the proposed APTCN. 
Fig. \ref{fig:aptcn_diag}(a) is the overview of the APTCN, which takes the feature $\mathcal{P}(\mathbf{x})$ as input, and processes it through a channel pruning block before entering the Efficient Channel Attention (ECA) module for feature enhancement. The enhanced output, denoted as $\mathbf{o}$, is then passed to a TCN block for feature analysis. 

Fig. \ref{fig:aptcn_diag}(b) shows a detailed visualisation of a $2$-layer TCN with kernel size $\mathbb{K}=3$. The first layer has a dilation factor $d=1$, which results in the conventional convolution operation. In subsequent upper layers with higher dilation factor $d$, a fixed gap is introduced between the inputs to account for the next available element in the sequence. Thus, the higher layers can retain the memory of the previous steps during the convolutions.

Regarding the channel pruning block of $\mathcal{P}(\mathbf{x})$, which reduces the channel sizes from $\mathcal{C}_{\text{in}}$ to $\mathcal{C}_{\text{out}}$; without loss of generality, we have set $\mathcal{C}_{\text{out}}=5$ in this paper. The pruned output, obtained by applying a channel pruning function $\mathcal{F}$ to $\mathcal{P}(\mathbf{x})$, is denoted as $\mathcal{P}^\prime(\mathbf{x}) = \mathcal{F}(\mathcal{P}(\mathbf{x}))$.

The enhanced version of $\mathcal{P}^\prime(\mathbf{x})$, denoted as $\mathbf{o}$, will then be obtained by using the Efficient Channel Attention (ECA) module, as it is a lightweight and efficient attention mechanism relative to other channel-attention approaches. 

\begin{equation}
\mathbf{a} = \sigma(\text{C1D}_\mathbb{K}(\text{GAP}(\mathcal{P}^\prime(\mathbf{x}))))\label{eq:eca}
\end{equation}

Equation (\ref{eq:eca}) presents the formulation of the ECA module, where the attention score $\mathbf{a}$ is computed by first obtaining a channel-wise view of the input using Global Average Pooling (GAP), then applying a 1D-CNN with kernel size $\mathbb{K}$ followed by a non-linear mapping function to generate the attention scores. The enhanced feature output $\mathbf{o}$ can then be obtained by multiplying the input $\mathcal{P}^\prime(\mathbf{x})$ with the score $\mathbf{a}$.

Finally, the TCN module takes feature $\mathbf{o}$ and generates the output $\hat{y}_T$, which is then aggregated in the \textit{Fusion} layer for classification. 
Note that the APTCN can adjust the depth by selecting a larger kernel size $\mathbb{K}$ and/or a deeper depth $N$ when a larger receptive field is needed. Each layer also has a residual connection to enhance the network performance. The memory of the current layer is $(\mathbb{K}-1)d$, and $d$ increases exponentially with the depth of the network, such that $d = O(2^i)$ at level $i \in N$ of the network.

\subsection{LSTM and C-FCN modules}

In addition to the probabilistic module, the proposed WPRCN integrates two specific modules for MTSC: an LSTM Network and a Causal-Fully Convolutional Network (C-FCN). Note that while various neural network architectures could be used along with the proposed probabilistic module for temporal data analysis, we have chosen the LSTM and C-FCN modules as instantiations to assess effectiveness in MTSC.

Specifically, an LSTM with a dropout layer is employed to extract temporal features from MTS data (refer to the middle module in Fig. \ref{fig:system_diag}). The architecture of the C-FCN module is depicted in the bottom Fig. \ref{fig:system_diag}, where the focus is primarily on the Squeeze-and-Excitation (SE) module, which investigates the cross-channel interactions after processing through Causal 1D-CNN (C-Conv1D), batch normalisation (BN), and \textit{ReLU}, leading up to the SE module.

The SE module differs from the ECA module in how it explores the channel relationships: it takes an MTS sequence $\mathbf{x} \in \mathbb{R}^{n, L}$, where $n$ is the number of features and $L$ is the sequence length, and computes a score $\omega = \sigma(\mathbf{W}_2 \text{ReLU}(\mathbf{W}_1 \text{GAP}(\mathbf{x})))$, where $\sigma$ is a Sigmoid function, $\mathbf{x}$ is the input, $\text{GAP}(\cdot)$ is the channel-wise global average pooling (GAP) operation, $\mathbf{W}_{1,2}$ are the factors that squeeze or excite the input, respectively. 

Note that the proposed WPRCN uses causal convolution in both C-FCN and APTCN modules, which prevents information leakage when modelling the sequential data with 1D-CNN. Finally, the aggregated features from the LSTM, C-FCN, and the probabilistic modules are concatenated and transformed at the \textit{Fusion} layer, and a Softmax layer is used for classification.

\section{Experiments}\label{sec:result}

The performance of the proposed WPRCN is benchmarked using the MTS data repository UEA from \cite{Bagnall2018The2018}, which covers $30$ datasets from categories of Human Activities, Electrocardiogram (ECG), Electroencephalogram (EEG), Magnetoencephalography (MAG), Audio, and others. The size of each dataset varies from $[27, 50000]$, with the number of features $n$ in the range of $[2, 1345]$ and sequence length $L$ in the range of $[8, 17984]$. Note that the data is normalised in $[0,1]^n$.

The performance of the proposed method is compared with seven benchmark approaches, i.e., 
\textit{DTW-1NN}, 
\textit{LSTM-FCN}, 
\textit{MLSTM-FCN}, 
\textit{Tap-Net}, 
\textit{CMFM-RF}, 
\textit{OS-CNN}, 
\textit{DA-Net} 
\cite{Shokoohi-Yekta2015OnCase, 
Karim2018LSTMClassification,
Karim2019MultivariateClassification,
Zhang2020TapNet:Network,
Baldan2021MultivariateRepresentation,
Tang2022Omni-ScaleClassification,
Chen2022DA-Net:Classification},
covering conventional machine learning approach and deep learning approaches using RNNs, CNNs and attention mechanisms.

The learning rate we used in the paper is searched within the range of $[1e-5, 1e-1]$ using the cyclical search proposed in \cite{Smith2017CyclicalNetworks}. For the key hyperparameters of the APTCN module and the loss function of WPRCN, we follow the settings in \cite{Bai2018AnModeling}: the kernel size $\mathbb{K}$ is searched from $\{3,5,7,11\}$ and with the depth of the network is defined in the range of $[3, 8]$. By varying $\mathbb{K}$ and the depth, APTCN can create variable receptive fields that target different datasets. We also set the attention key size as $\{1,3,5\}$ which considers different numbers of channels at the same time. The APTCN channel output is chosen between 20 and 25, with the dropout ratio set to $0.2$. As for the settings of the LSTM and the C-FCN, we follow the configuration in \cite{Karim2019MultivariateClassification}. For the result evaluation, we compute the classification Accuracy, Average Rank, and the number of Win/Tie counts as the evaluation metrics.

\subsection{Classification Performance Analysis}

\afterpage{%
  \clearpage
\begin{landscape}
\begin{table*}[!htp]
\centering
\caption{Performance comparison with the benchmark algorithms using 30 datasets from UEA data repository.}
\label{tbl:data_summary}
\scalebox{0.83}
{
\begin{tabular}{lcccccccc}
\hline\hline
                          \multicolumn{1}{c}{\textbf{Dataset}} & \textbf{WPRCN} & \textbf{\begin{tabular}[c]{@{}c@{}}DTW-1NND\\ \cite{Shokoohi-Yekta2015OnCase}\end{tabular}} & \textbf{\begin{tabular}[c]{@{}c@{}}LSTM-FCN\\ \cite{Karim2018LSTMClassification}\end{tabular}} & \textbf{\begin{tabular}[c]{@{}c@{}}MLSTM-FCN\\ \cite{Karim2019MultivariateClassification}\end{tabular}} & \textbf{\begin{tabular}[c]{@{}c@{}}TapNet\\ \cite{Zhang2020TapNet:Network}\end{tabular}} & \textbf{\begin{tabular}[c]{@{}c@{}}CMFM-RF\\ \cite{Baldan2021MultivariateRepresentation}\end{tabular}} & \textbf{\begin{tabular}[c]{@{}c@{}}OS-CNN\\ \cite{Tang2022Omni-ScaleClassification}\end{tabular}} & \textbf{\begin{tabular}[c]{@{}c@{}}DA-Net\\ \cite{Chen2022DA-Net:Classification}\end{tabular}}
                          \\ \hline
ArticularyWordRecognition (AR)            & \textbf{100.0} & 98.7                    & 98.5              & 98.6               & 98.7            & 98.8             & 98.8            & 98.0            \\
AtrialFibrillation (AF)                  & \textbf{62.5}  & 22.0                    & 23.3              & 20.0               & 33.3            & 19.1             & 23.3            & 46.7            \\
BasicMotions (BM)                         & 96.9           & 97.5                    & 93.1              & \textbf{100.0}     & \textbf{100.0}  & 98.2             & \textbf{100.0}  & 92.5            \\
CharacterTrajectories (CT)                & 99.1           & 98.9                    & 6.2               & 99.3               & 99.7            & 97.0             & \textbf{99.8}   & \textbf{99.8}   \\
Cricket (CR)                             & 96.9           & \textbf{100.0}          & 95.4              & 98.6               & 95.8            & 97.7             & 99.3            & 86.1            \\
DuckDuckGeese (DG)                       & \textbf{75.0}  & 60.0                    & 47.5              & 67.5               & 57.5            & 51.0             & 54.0            & 52.0            \\
ERing (ER)                               & \textbf{96.5}  & 13.3                    & 93.9              & 13.3               & 13.3            & 93.1             & 88.1            & 33.8            \\
EigenWorms (EW)                          & \textbf{89.8}  & 61.8                    & 80.4              & 80.9               & 48.9            & 89.5             & 41.4            & 48.9            \\
Epilepsy (EP)                            & \textbf{100.0} & 96.4                    & 94.9              & 96.4               & 97.1            & 99.9             & 98.0            & 83.3            \\
EthanolConcentration (EC)                 & 44.9           & 32.3                    & 26.7              & 27.4               & 32.3            & 26.0             & 24.0            & \textbf{87.4}   \\
FaceDetection (FD)                        & 59.0           & 52.9                    & 54.6              & 55.5               & 55.6            & 55.7             & 57.5            & \textbf{64.8}   \\
FingerMovements (FM)                     & \textbf{65.6}  & 53.0                    & 48.9              & 61.0               & 53.0            & 50.1             & 56.8            & 51.0            \\
HandMovementDirection (HM)               & \textbf{46.9}  & 23.1                    & 27.5              & 37.8               & 37.8            & 24.5             & 44.3            & 36.5            \\
Handwriting (HW)                         & 26.3           & 28.6                    & 22.0              & 54.7               & 35.7            & 27.4             & \textbf{66.8}   & 15.9            \\
Heartbeat (HB)                           & 74.2           & 71.7                    & 61.3              & 71.4               & 75.1            & \textbf{76.8}    & 48.9            & 62.4            \\
InsectWingbeat (IW)                      & 43.2           & -                       & 10.1              & 10.5               & 20.8            & \textbf{67.7}    & 66.7            & 56.7            \\
JapaneseVowels (JV)                      & 85.2           & 94.9                    & 9.8               & \textbf{99.2}      & 96.5            & 83.7             & 99.1            & 93.8            \\
LSST (LS)                                & 65.3           & 55.1                    & 60.9              & 64.6               & 56.8            & \textbf{67.3}    & 41.3            & 56.0            \\
Libras (LB)                              & 93.8           & 87.0                    & 91.3              & 92.2               & 85.0            & 84.7             & \textbf{95.0}   & 80.0            \\
MotorImagery (MI)                        & \textbf{75.0}  & 50.0                    & 54.9              & 53.0               & 59.0            & 50.3             & 53.5            & 50.0            \\
NATOPS (NA)                              & 93.8           & 88.3                    & 88.8              & 96.1               & 93.9            & 83.5             & \textbf{96.8}   & 87.8            \\
PEMS-SF (PE)                             & 88.3           & 71.1                    & 77.5              & 65.3               & 75.1            & \textbf{99.9}    & 76.0            & 86.7            \\
PenDigits (PD)                           & \textbf{99.3}  & 97.7                    & 98.8              & 98.7               & 98.0            & 95.2             & 98.5            & 98.0            \\
PhonemeSpectra (PS)                      & \textbf{32.8}  & 15.1                    & 24.4              & 27.5               & 17.5            & 28.4             & 29.9            & 9.3             \\
RacketSports (RS)                        & 85.2           & 80.3                    & 83.0              & \textbf{88.2}      & 86.8            & 80.6             & 87.7            & 80.3            \\
SelfRegulationSCP1 (SR1)                   & 84.4           & 77.5                    & 70.6              & 86.7               & 65.2            & 82.0             & 83.5            & \textbf{92.4}   \\
SelfRegulationSCP2 (SR2)                  & \textbf{64.8}  & 53.9                    & 51.9              & 52.2               & 55.0            & 41.8             & 53.2            & 56.1            \\
SpokenArabicDigits (SAD)                   & \textbf{99.9}  & 96.3                    & 9.2               & 99.4               & 98.3            & 97.6             & 99.7            & 98.0            \\
StandWalkJump (SWJ)                        & \textbf{75.0}  & 20.0                    & 39.2              & 46.7               & 40.0            & 36.3             & 38.3            & 40.0            \\
UWaveGestureLibrary (UL)                 & 87.8           & 90.3                    & 84.3              & 85.7               & 89.4            & 77.5             & \textbf{92.7}   & 83.3            \\
                                     &                &                         &                   &                    &                 &                  &                 &                 \\
Avg Accuracy                         & \textbf{76.9}  & 65.1                    & 57.6              & 68.3               & 65.7            & 69.4             & 70.4            & 67.6            \\
Avg Rank                             & \textbf{2.5}   & 5.5                     & 5.8               & 3.9                & 4.2             & 5.1              & 3.6             & 5.3             \\
Win/Ties                             & \textbf{14}    & 1                       & 0                 & 3                  & 1               & 4                & 6               & 4              \\
\hline\hline
\end{tabular}
}
\end{table*}
\end{landscape}
\clearpage
}

The classification results of the proposed WRPCN and the benchmark algorithms are presented in Table \ref{tbl:data_summary}, where 30 UEA datasets from different fields with various sizes, lengths, and feature dimensions are used. The best result for each dataset is highlighted in bold font for clarity. Note that "-" in the table indicates the algorithm fails to produce the results.

The overall average accuracy of the proposed WPRCN is the best, which has $7$\% higher average accuracy than the best benchmark algorithm \textit{OS-CNN}. Notably, the WPRCN reports superior classification accuracy across a wide range of sequence lengths ($L = [8, 3000]$), input features ($n = [2, 1345]$), and most importantly, it maintains outstanding classification performance even when limited training data is available, 
i.e., for datasets \textit{AR}, \textit{ER}, and \textit{SWJ}, they only have training data sizes that vary from $12$ to $30$. The proposed approach achieves at least $16$\%, $8$\%, and $28$\% higher classification accuracy than the best benchmark algorithms on these datasets, respectively.

This is evidence of the enhancement introduced by the AWPG, as the $j_0$ and $m$ can be adjusted according to different amounts of data. Additionally, the WPRCN reports the top rank and achieves the highest number of Win/Tie counts.

Note that the WPRCN shows a robust classification performance across a wide range of physiological datasets (e.g., \textit{AtrialFibrillation (ECG)}, \textit{FingerMovements (EEG)}, \textit{HandMovementDirection (MEG)}). For such physiological data classification tasks, the presence of various perturbations during data acquisition and inherent non-stationarity introduce significant challenges, rendering classification performance highly contingent upon the specific application scenarios \cite{AlZoubi2015AffectClassifiers}. The proposed WPRCN utilises the probabilistic feature module (i.e., the AWPG and the APTCN) to capture and handle such perturbations and non-stationaries in the data, enabling the generation of corresponding probabilistic features that substantially enhance the classification performance. 

\begin{figure}[!htp]
\centering
\includegraphics[trim = 0mm 0mm 0mm 0mm,clip,width=0.91\linewidth]{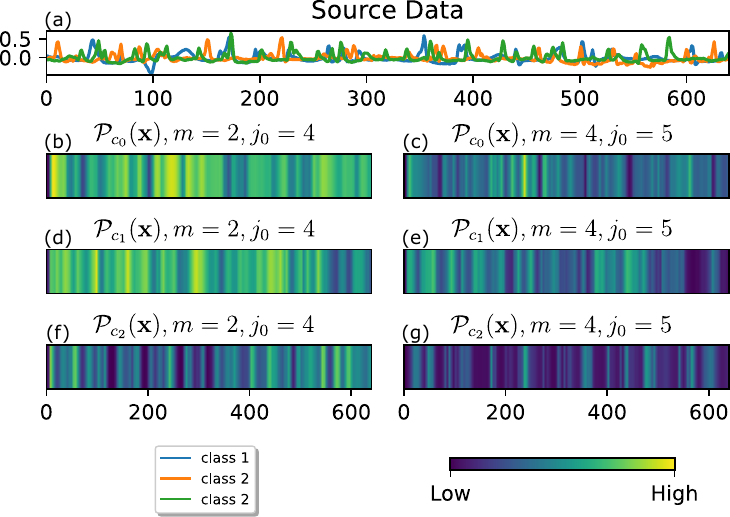}
\centering
\caption{The probabilistic features $\mathcal{P}(\mathbf{x})_{m,j_0}$ for three different classes. (a) The first channel of the source signal for dataset \textit{AtrialFibrillation}, which is a three-class dataset with $L=640$, $n=2$. (b-g) The $\mathcal{P}_{c}(\mathbf{x})_{m,j_0}$ for different $m$ and $j_0$ are presented. Each row shows the same class of data with different $m$ and $j_0$, and each column shows the same $m$ and $j_0$ across different classes of data. 
}\label{fig:distinct_pdf}
\end{figure}

Fig. \ref{fig:distinct_pdf} shows the visualisation of the physiological data from dataset \textit{AtrialFibrillation}, where the first channel of the dataset is shown in (a), the feature with different $m$ and $j_0$ for different classes are shown in (b-g). From the figure, the fluctuations of the signal are well captured by the AWPG module, which is then analysed by the APTCN module to extract features from different levels of information for classification. By comparing the $\mathcal{P}_{c}(\mathbf{x})_{m,j_0}$ for class $c$ using different $m$ and $j_0$ in each row of Fig. \ref{fig:distinct_pdf}, different levels of information positioned in different timestamps are revealed. For example, as illustrated in subfigures (b) and (c), the latter contains more fluctuations due to a higher order of $m$ and $j_0$ being used. Therefore, by aggregating all the combinations of $m$ and $j_0$, the resulting $\mathcal{P}(\mathbf{x})$ encapsulates the comprehensive information essential for the temporal data modelling tasks.

\begin{figure}[htp]
\centering
\includegraphics[trim = 0mm 0mm 0mm 0mm,clip,width=\linewidth]{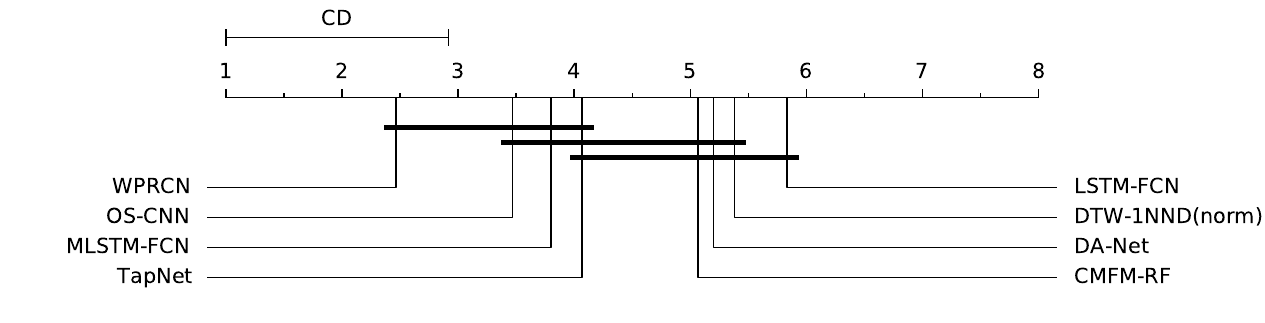}
\caption{CD diagram of all the algorithms using $30$ UEA datasets.} \label{fig:cd_diag}
\end{figure}

We further show the critical difference (CD) diagram in Fig. \ref{fig:cd_diag}, which evaluates the statistical significance of the performance difference across all the algorithms. It is clear to see that the proposed WPRCN reports the best ranking against all the benchmark algorithms.

\subsection{Further Analysis}

Although WPRCN reports the best classification performance against all the other algorithms, 
there exist some cases in which the performance of WPRCN lags. We selected datasets \{\textit{EC}, \textit{FD}, \textit{HW}, \textit{IW}, \textit{JV}, \textit{PE}, \textit{SR1}, and \textit{UL}\} for discussion, as the accuracy of WPRCN deviates by more than $5\%$ when compared to the best benchmark algorithm. Four benchmark algorithms are involved, i.e., \textit{DA-Net}, \textit{OS-CNN}, \textit{CMFM-RF}, and \textit{MLSTM-FCN}.

For \textit{DA-Net}, it reports top accuracy on datasets \textit{EC, FD,} and \textit{SR1}. This stems from the design of \textit{DA-Net}, which utilises dual attention mechanisms to capture both global and local features. In this sense, for datasets with longer sequence lengths (such as datasets \textit{EC} and \textit{SR1} have the sequence length of $1751$ and $896$, respectively), a higher classification accuracy can be observed. However, despite \textit{DA-Net}'s effectiveness on these three datasets, its generalisability to diverse data structures appears to be limited (see the results in Table \ref{tbl:data_summary} and Fig. \ref{fig:cd_diag}).

\textit{OS-CNN} outperforms on datasets \textit{HW} and \textit{UL}. These two datasets have a commonality by having the same number of input features ($n=2$), with $8$ and $26$ different classes, respectively. 
WPRCN shows less optimal results on these two datasets as we only fix the APTCN channel output size to $20$ or $25$, which streamlines the hyper-parameter tuning process. This configuration, while expediting model tuning time, suggests that future refinements can be explored for datasets with specific characteristics such as data with more classes but limited training data.

For \textit{CMFM-RF}, it shows the best performance on datasets \textit{IW} and  \textit{PE}. 
As discussed in \textit{CMFM-RF}, these two datasets can be perfectly represented by manually selected features, and therefore,  \textit{CMFM-RF} reports better accuracy \cite{Baldan2021MultivariateRepresentation}. However, for datasets that cannot be adequately characterised by the features designed in \textit{CMFM-RF}, the proposed WPRCN, with its powerful feature extraction capability, reports superior performance. Moreover, it is worth integrating the proposed probabilistic module with \textit{CMFM-RF}, as the distinctive features can improve the performance of the classifiers.

For dataset \textit{JV}, \textit{MLSTM-FCN} reports better results than WPRCN. This is due to the data normalisation process employed in the data preprocessing stage of WPRCN. Specifically, the dataset \textit{JV}, with its varying data lengths, has been standardised the length to $29$ using zero padding. However, this process, combined with the normalisation, introduces distortions that affect the performance of the three feature extraction modules, as the sequences are slightly altered.

Finally, there are two datasets, i.e., \textit{HM} and \textit{PS} that WPRCN yields the best results, but the absolute accuracies for these two datasets remain low. This is due to the inter-class similarities in the datasets, which affects the feature extraction and classification capabilities in all the algorithms. It is worth integrating the proposed probabilistic module with other techniques, such as temporal attention or contrastive learning, to enhance the classification performance,e as this module is designed to be seamlessly integrated with other frameworks.

\clearpage

\subsection{Ablation Tests}\label{sec:ablation}

Previous results demonstrate the outstanding classification performance of the proposed WPRCN. This performance can be attributed to: (i) the incorporation of powerful and informative probabilistic feature generation and analysis modules, and (ii) the channel attention modules that learn the channel weights, thereby improving the classification results. In this section, we perform three ablation tests to show the importance of these modules, and the results are given in Table \ref{tbl:ablation}.

In Table \ref{tbl:ablation}, the tests labelled as \textit{A1, A2,} and \textit{A3} correspond to three distinct ablation tests: (i) \textit{A1}, excludes the entire probabilistic module, where both the probabilistic feature generator (AWPG) and its associated analyser (APTCN) are removed, (ii) \textit{A2}, only omits the probabilistic feature generator (AWPG); and (iii) \textit{A3}, removes the ECA module in APTCN. 

Table \ref{tbl:ablation} reveals the performance rank, with a noticeable progression in average accuracy from \textit{A1, A2, A3}, followed by WPRCN. This trend reflects the growing contribution of the components from the probabilistic module. Specifically, the removal of the entire probabilistic module (as in Test \textit{A1}) results in a reduction of $6.1\%$ in average accuracy. Excluding only the probabilistic feature generator (as in Test \textit{A2}) reports a $5.8\%$ drop, and omitting the attention module (as in Test \textit{A3}) leads to a $4.3\%$ decrease in average accuracy. The presented metrics, i.e., the average accuracy, average rank, and win/tie counts, emphasise an integral part of each component in the probabilistic module, as their omission compromises the classification performance.

\begin{figure}[htp]
\centering
\includegraphics[trim = 0mm 0mm 0mm 0mm,clip,width=0.94\linewidth]{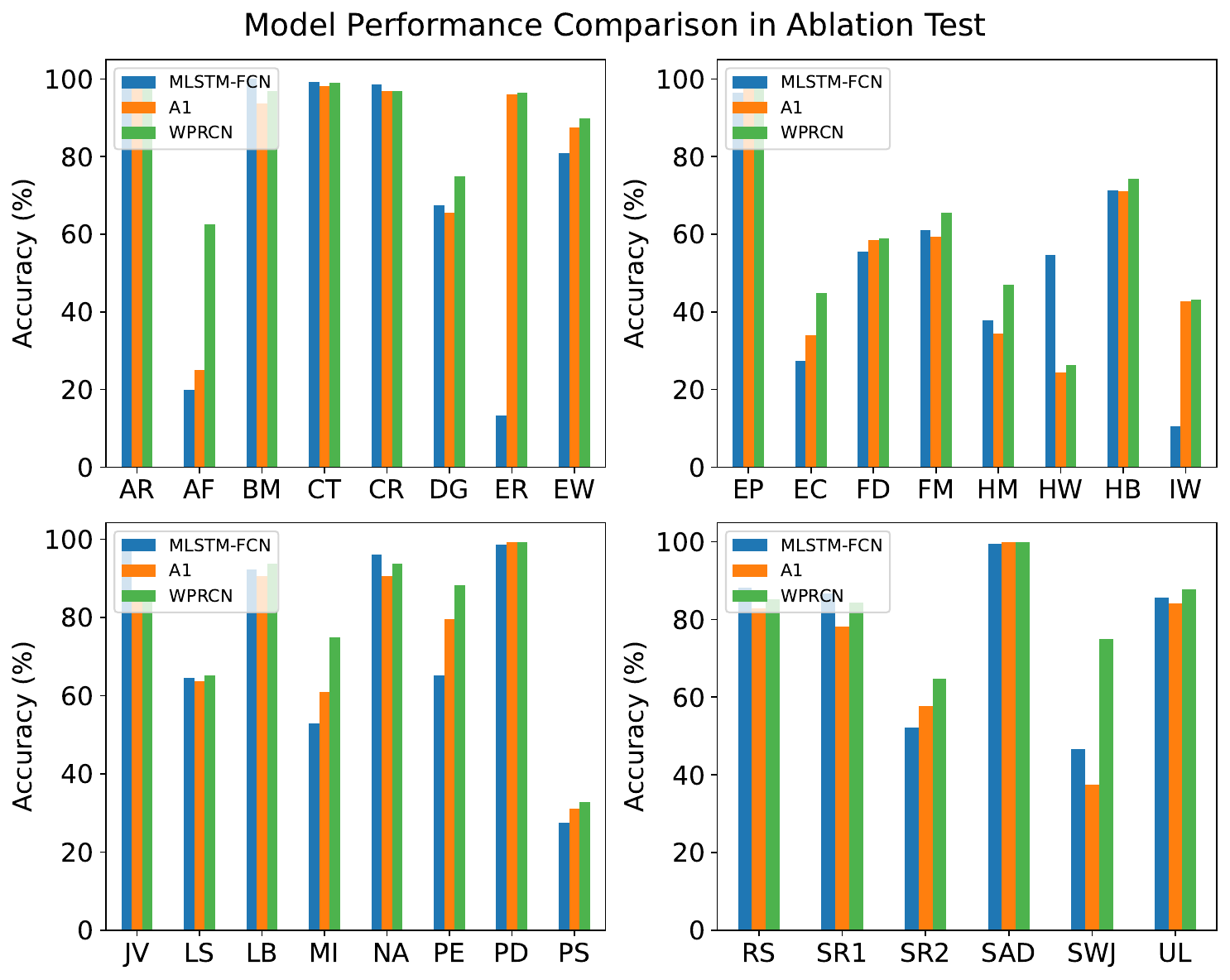}
\caption{Performance comparison among \textit{MLSTM-FCN}, \textit{Test A1}, and WPRCN.} \label{fig:ablation_3_tests}
\end{figure}

We further note that \textit{Test A1} has a similar model architecture as \textit{MLSTM-FCN}, so we present a detailed performance comparison among \textit{MLSTM-FCN, Test A1}, and WPRCN in Fig \ref{fig:ablation_3_tests}. 
Based on Table \ref{tbl:data_summary}-\ref{tbl:ablation} and 
Fig. \ref{fig:ablation_3_tests}, we would expect both \textit{MLSTM-FCN} and \textit{Test A1} to exhibit similar performances. 
However, there are some exceptions where deviations in accuracy can be observed, such as datasets \textit{ER, HW, IW, SR1}. For \textit{MLSTM-FCN}, it has $30\%$ and $9\%$ higher accuracy than \textit{Test A1} on datasets \textit{HW, SR1}, respectively. But it also reports $83\%$ and $32\%$ lower accuracy than \textit{Test A1} in \textit{ER, IW}. 
This is attributed to the dimensional shuffling operation present within the LSTM module in \textit{MLSTM-FCN}, which significantly impacts sequential modelling. Consider, for example, the dataset \textit{IW}: it contains more features ($n=200$) than the sequence length ($L=22$), therefore, after the dimension shuffling operation in \textit{MLSTM-FCN}, the LSTM prioritises modelling the feature channel over the temporal channels, thereby losing the ability to model the temporal dependencies.

\begin{table}[!htp]
\centering
\caption{The ablation test results, where the \textit{Test A1} removes the entire probabilistic module, i.e., AWPG and APTCN; \textit{Test  A2} removes the probabilistic feature generator AWPG so that APTCN will analyse the source time series data; \textit{Test A3} removes the channel attention mechanism. }
\label{tbl:ablation}
\scalebox{0.89}
{
\begin{tabular}{lcccc}
\hline \hline
                          & A1            & A2            & A3            & WPRCN          \\ \hline
ArticularyWordRecognition (AR) & 99.2          & 98.8          & 99.2          & \textbf{100.0} \\
AtrialFibrillation (AF)        & 25.0          & 50.0          & \textbf{62.5} & \textbf{62.5}  \\
BasicMotions (BM)             & 93.8          & 93.8          & 93.8          & \textbf{96.9}  \\
CharacterTrajectories (CT)     & 98.2          & 98.4          & 98.4          & \textbf{99.1}  \\
Cricket (CR)                  & \textbf{96.9} & \textbf{96.9} & \textbf{96.9} & \textbf{96.9}  \\
DuckDuckGeese (DG)            & 65.6          & 68.8          & 71.9          & \textbf{75.0}  \\
ERing (ER)                    & 96.1          & 88.7          & 93.8          & \textbf{96.5}  \\
EigenWorms (EW)               & 87.5          & 85.2          & 85.2          & \textbf{89.8}  \\
Epilepsy (EP)                 & 97.7          & 97.7          & 97.7          & \textbf{100.0} \\
EthanolConcentration (EC)      & 34.0          & 32.0          & 30.9          & \textbf{44.9}  \\
FaceDetection (FD)            & 58.5          & 54.2          & 55.4          & \textbf{59.0}  \\
FingerMovements (FM)          & 59.4          & 56.3          & 57.8          & \textbf{65.6}  \\
HandMovementDirection (HM)    & 34.4          & 26.6          & 43.8          & \textbf{46.9}  \\
Handwriting (HW)              & 24.3          & 23.3          & 25.8          & \textbf{26.3}  \\
Heartbeat (HB)                & 71.1          & 72.7          & 72.7          & \textbf{74.2}  \\
InsectWingbeat (IW)           & 42.7          & 39.7          & 42.0          & \textbf{43.2}  \\
JapaneseVowels (JV)            & 84.0          & 80.5          & 84.0          & \textbf{85.2}  \\
LSST (LS)                      & 63.8          & 64.7          & \textbf{65.3} & \textbf{65.3}  \\
Libras (LB)                   & 90.6          & 93.0          & 93.0          & \textbf{93.8}  \\
MotorImagery (MI)             & 60.9          & 64.1          & 60.9          & \textbf{75.0}  \\
NATOPS (NA)                    & 90.6          & 89.8          & 90.6          & \textbf{93.8}  \\
PEMS-SF (PE)                  & 79.7          & 82.8          & 86.7          & \textbf{88.3}  \\
PenDigits (PD)                & 99.2          & 99.2          & \textbf{99.3} & \textbf{99.3}  \\
PhonemeSpectra (PS)           & 31.2          & 26.2          & 29.1          & \textbf{32.8}  \\
RacketSports (RS)             & 82.8          & 77.3          & 77.3          & \textbf{85.2}  \\
SelfRegulationSCP1 (SR1)       & 78.1          & 73.4          & 82.0          & \textbf{84.4}  \\
SelfRegulationSCP2 (SR2)       & 57.8          & 60.9          & 60.9          & \textbf{64.8}  \\
SpokenArabicDigits (SAD)       & \textbf{99.9} & 99.4          & \textbf{99.9} & \textbf{99.9}  \\
StandWalkJump (SWJ)            & 37.5          & 62.5          & 37.5          & \textbf{75.0}  \\
UWaveGestureLibrary (UL)      & 84.1          & 76.2          & 84.4          & \textbf{87.8}  \\
                          &               &               &               &                \\
Avg Accuracy              & 70.8          & 71.1          & 72.6          & \textbf{76.9}  \\
Avg Rank                  & 3             & 3.3           & 2.6           & \textbf{1.1}   \\
Win/Ties                  & 2             & 1             & 5             & \textbf{30}   \\
\hline \hline
\end{tabular}
}
\end{table}

\clearpage

\section{Conclusion}\label{sec:final_remark}

In this paper, we propose a novel Wavelet Probabilistic Recurrent Convolutional Network (WPRCN) for multivariate time series classification, which demonstrates particular effectiveness in handling non-stationary environments, data scarcity, and perturbations. The proposed WPRCN consists of three modules, each considering a distinct temporal feature analysis approach: the LSTM, the C-FCN, and most importantly, the probabilistic module that generates distinctive probabilistic features for each class of data, thereby enabling the model to capture key insights for delineating the class boundaries in MTSC tasks.

The AWPG module can adaptively select the most suitable probabilistic model, and generate probabilistic features with different smoothness (caused by the inherent characteristic of the data itself and the size of available data). The enhanced probabilistic features are learned in the APTCN module via a channel attention block and a TCN block. Finally, a \textit{Fusion} layer aggregates the temporal features from the three different modules and performs classification using the combined features.

By introducing the probabilistic module (AWPG and APTCN), we extend the application of wavelet probabilistic neural networks to deep neural networks for MTSC. The \textit{Adaptive Network} employed in AWPG can adaptively select the optimal views without the need for human intervention. The proposed probabilistic module significantly enhances the performance of the MTSC tasks, especially for data in non-stationary environments, with limited training data being available, and possibly having more perturbations (e.g., physiological data). 
The proposed WPRCN has been evaluated with 30 different MTS datasets covering various fields of applications and reports the best performance in all the metrics. Moreover, the ablation study confirms the importance of the proposed probabilistic module.

The selected instantiation of the proposed WPRCN demonstrates the seamless integration capability of the proposed probabilistic module for time series analysis.
Future research could extend this paradigm by incorporating the probabilistic module into alternative neural network frameworks for temporal data analysis.
Moreover, currently, WPRCN only considers channel attention for feature enhancement, future research can incorporate other attention mechanisms, such as temporal attention, to learn more meaningful features for MTSC.

\newpage

\vfill

\bibliographystyle{elsarticle-num}
\bibliography{main}

\begin{thebibliography}{10}
\expandafter\ifx\csname url\endcsname\relax
  \def\url#1{\texttt{#1}}\fi
\expandafter\ifx\csname urlprefix\endcsname\relax\def\urlprefix{URL }\fi
\expandafter\ifx\csname href\endcsname\relax
  \def\href#1#2{#2} \def\path#1{#1}\fi

\bibitem{Sezer2020Financial20052019}
O.~B. Sezer, M.~U. Gudelek, A.~M. Ozbayoglu, {Financial time series forecasting with deep learning : A systematic literature review: 2005–2019}, Applied Soft Computing 90 (2020) 106181.
\newblock \href {https://doi.org/10.1016/j.asoc.2020.106181} {\path{doi:10.1016/j.asoc.2020.106181}}.

\bibitem{Barria2011DetectionVariables}
J.~A. Barria, S.~Thajchayapong, {Detection and Classification of Traffic Anomalies Using Microscopic Traffic Variables}, IEEE Transactions on Intelligent Transportation Systems 12~(3) (2011) 695--704.
\newblock \href {https://doi.org/10.1109/TITS.2011.2157689} {\path{doi:10.1109/TITS.2011.2157689}}.

\bibitem{Ingolfsson2021ECG-TCN:Network}
T.~M. Ingolfsson, X.~Wang, M.~Hersche, A.~Burrello, L.~Cavigelli, L.~Benini, {ECG-TCN: Wearable Cardiac Arrhythmia Detection with a Temporal Convolutional Network}, in: 2021 IEEE 3rd International Conference on Artificial Intelligence Circuits and Systems (AICAS), IEEE, 2021, pp. 1--4.
\newblock \href {https://doi.org/10.1109/AICAS51828.2021.9458520} {\path{doi:10.1109/AICAS51828.2021.9458520}}.

\bibitem{IsmailFawaz2019DeepReview}
H.~Ismail~Fawaz, G.~Forestier, J.~Weber, L.~Idoumghar, P.~A. Muller, {Deep learning for time series classification: a review}, Data Mining and Knowledge Discovery 33~(4) (2019) 917--963.
\newblock \href {https://doi.org/10.1007/s10618-019-00619-1} {\path{doi:10.1007/s10618-019-00619-1}}.

\bibitem{Sutskever2014SequenceNetworks}
I.~Sutskever, O.~Vinyals, Q.~V. Le, {Sequence to sequence learning with neural networks}, Advances in Neural Information Processing Systems 4~(January) (2014) 3104--3112.

\bibitem{He2016DeepRecognition}
K.~He, X.~Zhang, S.~Ren, J.~Sun, {Deep Residual Learning for Image Recognition}, in: 2016 IEEE Conference on Computer Vision and Pattern Recognition (CVPR), IEEE, 2016, pp. 770--778.
\newblock \href {https://doi.org/10.1109/CVPR.2016.90} {\path{doi:10.1109/CVPR.2016.90}}.

\bibitem{Sajjad2020AForecasting}
M.~Sajjad, Z.~A. Khan, A.~Ullah, T.~Hussain, W.~Ullah, M.~Y. Lee, S.~W. Baik, {A Novel CNN-GRU-Based Hybrid Approach for Short-Term Residential Load Forecasting}, IEEE Access 8 (2020) 143759--143768.
\newblock \href {https://doi.org/10.1109/ACCESS.2020.3009537} {\path{doi:10.1109/ACCESS.2020.3009537}}.

\bibitem{Wang2019DevelopmentSurvey}
W.~Wang, Y.~Yang, {Development of convolutional neural network and its application in image classification: a survey}, Optical Engineering 58~(04) (2019) 1.
\newblock \href {https://doi.org/10.1117/1.OE.58.4.040901} {\path{doi:10.1117/1.OE.58.4.040901}}.

\bibitem{Wang2017TimeBaseline}
Z.~Wang, W.~Yan, T.~Oates, {Time series classification from scratch with deep neural networks: A strong baseline}, in: 2017 International Joint Conference on Neural Networks (IJCNN), IEEE, 2017, pp. 1578--1585.
\newblock \href {https://doi.org/10.1109/IJCNN.2017.7966039} {\path{doi:10.1109/IJCNN.2017.7966039}}.

\bibitem{IsmailFawaz2020InceptionTime:Classification}
H.~Ismail~Fawaz, B.~Lucas, G.~Forestier, C.~Pelletier, D.~F. Schmidt, J.~Weber, G.~I. Webb, L.~Idoumghar, P.~A. Muller, F.~Petitjean, {InceptionTime: Finding AlexNet for time series classification}, Data Mining and Knowledge Discovery 34~(6) (2020) 1936--1962.
\newblock \href {https://doi.org/10.1007/s10618-020-00710-y} {\path{doi:10.1007/s10618-020-00710-y}}.

\bibitem{Karim2018LSTMClassification}
F.~Karim, S.~Majumdar, H.~Darabi, S.~Chen, {LSTM Fully Convolutional Networks for Time Series Classification}, IEEE Access 6 (2018) 1662--1669.
\newblock \href {https://doi.org/10.1109/ACCESS.2017.2779939} {\path{doi:10.1109/ACCESS.2017.2779939}}.

\bibitem{Karim2019MultivariateClassification}
F.~Karim, S.~Majumdar, H.~Darabi, S.~Harford, \href{https://doi.org/10.1016/j.neunet.2019.04.014}{{Multivariate LSTM-FCNs for time series classification}}, Neural Networks 116 (2019) 237--245.
\newblock \href {https://doi.org/10.1016/j.neunet.2019.04.014} {\path{doi:10.1016/j.neunet.2019.04.014}}.
\newline\urlprefix\url{https://doi.org/10.1016/j.neunet.2019.04.014}

\bibitem{Chaovalit2011DiscreteMining}
P.~Chaovalit, A.~Gangopadhyay, G.~Karabatis, Z.~Chen, {Discrete wavelet transform-based time series analysis and mining}, ACM Computing Surveys 43~(2) (2011) 1--37.
\newblock \href {https://doi.org/10.1145/1883612.1883613} {\path{doi:10.1145/1883612.1883613}}.

\bibitem{Zhang2004AClassification}
H.~Zhang, T.~B. Ho, M.~S. Lin, \href{https://link.springer.com/chapter/10.1007/978-3-540-24775-3_71}{{A Non-parametric Wavelet Feature Extractor for Time Series Classification}}, Lecture Notes in Computer Science (including subseries Lecture Notes in Artificial Intelligence and Lecture Notes in Bioinformatics) 3056 (2004) 595--603.
\newblock \href {https://doi.org/10.1007/978-3-540-24775-3{\_}71} {\path{doi:10.1007/978-3-540-24775-3{\_}71}}.
\newline\urlprefix\url{https://link.springer.com/chapter/10.1007/978-3-540-24775-3_71}

\bibitem{Wang2018MultilevelAnalysis}
J.~Wang, Z.~Wang, J.~Li, J.~Wu, \href{https://dl.acm.org/doi/10.1145/3219819.3220060}{{Multilevel wavelet decomposition network for interpretable time series analysis}}, Proceedings of the ACM SIGKDD International Conference on Knowledge Discovery and Data Mining (2018) 2437--2446\href {https://doi.org/10.1145/3219819.3220060} {\path{doi:10.1145/3219819.3220060}}.
\newline\urlprefix\url{https://dl.acm.org/doi/10.1145/3219819.3220060}

\bibitem{Pancholi2023SourceSignal}
S.~Pancholi, A.~Giri, A.~Jain, L.~Kumar, S.~Roy, {Source Aware Deep Learning Framework for Hand Kinematic Reconstruction Using EEG Signal}, IEEE Transactions on Cybernetics 53~(7) (2023) 4094--4106.
\newblock \href {https://doi.org/10.1109/TCYB.2022.3166604} {\path{doi:10.1109/TCYB.2022.3166604}}.

\bibitem{Li2020WaveletClassification}
Q.~Li, L.~Shen, S.~Guo, Z.~Lai, {Wavelet Integrated CNNs for Noise-Robust Image Classification}, Proceedings of the IEEE Computer Society Conference on Computer Vision and Pattern Recognition (2020) 7243--7252\href {https://doi.org/10.1109/CVPR42600.2020.00727} {\path{doi:10.1109/CVPR42600.2020.00727}}.

\bibitem{Garcia-Trevino2014StructuralClassification}
E.~S. Garc{\'{i}}a-Trevi{\~{n}}o, J.~A. Barria, {Structural generative descriptions for time series classification}, IEEE Transactions on Cybernetics 44~(10) (2014) 1978--1991.
\newblock \href {https://doi.org/10.1109/TCYB.2014.2322310} {\path{doi:10.1109/TCYB.2014.2322310}}.

\bibitem{Garcia-Trevino2024WaveletNetworks}
E.~S. Garcia-Trevino, P.~Yang, J.~A. Barria, {Wavelet Probabilistic Neural Networks}, IEEE Transactions on Neural Networks and Learning Systems 35~(1) (2024) 376--389.
\newblock \href {https://doi.org/10.1109/TNNLS.2022.3174705} {\path{doi:10.1109/TNNLS.2022.3174705}}.

\bibitem{AlZoubi2015AffectClassifiers}
O.~AlZoubi, D.~Fossati, S.~D’Mello, R.~A. Calvo, {Affect detection from non-stationary physiological data using ensemble classifiers}, Evolving Systems 6~(2) (2015) 79--92.
\newblock \href {https://doi.org/10.1007/s12530-014-9123-z} {\path{doi:10.1007/s12530-014-9123-z}}.

\bibitem{Chen2022DA-Net:Classification}
R.~Chen, X.~Yan, S.~Wang, G.~Xiao, {DA-Net: Dual-attention network for multivariate time series classification}, Information Sciences 610 (2022) 472--487.
\newblock \href {https://doi.org/10.1016/j.ins.2022.07.178} {\path{doi:10.1016/j.ins.2022.07.178}}.

\bibitem{HajizadehTahan2022DevelopmentClassification}
M.~Hajizadeh~Tahan, M.~Ghasemzadeh, S.~Asadi, {Development of fully convolutional neural networks based on discretization in time series classification}, IEEE Transactions on Knowledge and Data Engineering (2022) 1--1\href {https://doi.org/10.1109/tkde.2022.3177724} {\path{doi:10.1109/tkde.2022.3177724}}.

\bibitem{He2023Rel-CNN:Classification}
F.~He, T.~y. Fu, W.~C. Lee, {Rel-CNN: Learning Relationship Features in Time Series for Classification}, IEEE Transactions on Knowledge and Data Engineering 35~(7) (2023) 7412--7426.
\newblock \href {https://doi.org/10.1109/TKDE.2022.3186963} {\path{doi:10.1109/TKDE.2022.3186963}}.

\bibitem{Chen2021}
W.~Chen, K.~Shi, {Multi-scale Attention Convolutional Neural Network for time series classification}, Neural Networks 136 (2021) 126--140.
\newblock \href {https://doi.org/10.1016/j.neunet.2021.01.001} {\path{doi:10.1016/j.neunet.2021.01.001}}.

\bibitem{Yue2025}
J.~Yue, J.~Wang, S.~Zhang, Z.~Ma, Y.~Shi, Y.~Lin, {TV-Net: Temporal-Variable feature harmonizing Network for multivariate time series classification and interpretation}, Neural Networks 182 (2025).
\newblock \href {https://doi.org/10.1016/j.neunet.2024.106896} {\path{doi:10.1016/j.neunet.2024.106896}}.

\bibitem{Zhang2020TapNet:Network}
X.~Zhang, Y.~Gao, J.~Lin, C.~T. Lu, {TapNet: Multivariate time series classification with attentional prototypical network}, AAAI 2020 - 34th AAAI Conference on Artificial Intelligence (2020) 6845--6852\href {https://doi.org/10.1609/aaai.v34i04.6165} {\path{doi:10.1609/aaai.v34i04.6165}}.

\bibitem{Tang2022Omni-ScaleClassification}
W.~Tang, G.~Long, L.~Liu, T.~Zhou, M.~Blumenstein, J.~Jiang, {Omni-Scale Cnns: a Simple and Effective Kernel Size Configuration for Time Series Classification}, ICLR 2022 - 10th International Conference on Learning Representations (2022).

\bibitem{Tahan2023AClassification}
M.~H. Tahan, M.~Ghasemzadeh, S.~Asadi, {A Novel Embedded Discretization-Based Deep Learning Architecture for Multivariate Time Series Classification}, IEEE Transactions on Industrial Informatics 19~(4) (2023) 5976--5984.
\newblock \href {https://doi.org/10.1109/TII.2022.3188839} {\path{doi:10.1109/TII.2022.3188839}}.

\bibitem{Wu2022TimesNet:Analysis}
H.~Wu, T.~Hu, Y.~Liu, H.~Zhou, J.~Wang, M.~Long, \href{http://arxiv.org/abs/2210.02186}{{TimesNet: Temporal 2D-Variation Modeling for General Time Series Analysis}} (2022).
\newline\urlprefix\url{http://arxiv.org/abs/2210.02186}

\bibitem{Baldan2021MultivariateRepresentation}
F.~J. Bald{\'{a}}n, J.~M. Ben{\'{i}}tez, {Multivariate times series classification through an interpretable representation}, Information Sciences 569 (2021) 596--614.
\newblock \href {https://doi.org/10.1016/j.ins.2021.05.024} {\path{doi:10.1016/j.ins.2021.05.024}}.

\bibitem{Oord2016WaveNet:Audio}
A.~v.~d. Oord, S.~Dieleman, H.~Zen, K.~Simonyan, O.~Vinyals, A.~Graves, N.~Kalchbrenner, A.~Senior, K.~Kavukcuoglu, \href{https://arxiv.org/abs/1609.03499v2}{{WaveNet: A Generative Model for Raw Audio}} (9 2016).
\newline\urlprefix\url{https://arxiv.org/abs/1609.03499v2}

\bibitem{Bai2018AnModeling}
S.~Bai, J.~Z. Kolter, V.~Koltun, {An Empirical Evaluation of Generic Convolutional and Recurrent Networks for Sequence Modeling} (3 2018).

\bibitem{Brauwers2021ALearning}
G.~Brauwers, F.~Frasincar, {A General Survey on Attention Mechanisms in Deep Learning}, IEEE Transactions on Knowledge and Data Engineering (2021) 1--1\href {https://doi.org/10.1109/tkde.2021.3126456} {\path{doi:10.1109/tkde.2021.3126456}}.

\bibitem{Vaswani2017AttentionNeed}
A.~Vaswani, N.~Shazeer, N.~Parmar, J.~Uszkoreit, L.~Jones, A.~N. Gomez, {\L}.~Kaiser, I.~Polosukhin, {Attention is all you need}, Advances in Neural Information Processing Systems 2017-Decem (2017) 5999--6009.

\bibitem{Woo2018CBAM:Module}
S.~Woo, J.~Park, J.~Y. Lee, I.~S. Kweon, {CBAM: Convolutional block attention module}, Lecture Notes in Computer Science (including subseries Lecture Notes in Artificial Intelligence and Lecture Notes in Bioinformatics) 11211 LNCS (2018) 3--19.
\newblock \href {https://doi.org/10.1007/978-3-030-01234-2{\_}1} {\path{doi:10.1007/978-3-030-01234-2{\_}1}}.

\bibitem{Wang2020ECA-Net:Networks}
Q.~Wang, B.~Wu, P.~Zhu, P.~Li, W.~Zuo, Q.~Hu, {ECA-Net: Efficient channel attention for deep convolutional neural networks}, Proceedings of the IEEE Computer Society Conference on Computer Vision and Pattern Recognition (2020) 11531--11539\href {https://doi.org/10.1109/CVPR42600.2020.01155} {\path{doi:10.1109/CVPR42600.2020.01155}}.

\bibitem{Hu2020Squeeze-and-ExcitationNetworks}
J.~Hu, L.~Shen, S.~Albanie, G.~Sun, E.~Wu, {Squeeze-and-Excitation Networks}, IEEE Transactions on Pattern Analysis and Machine Intelligence 42~(8) (2020) 2011--2023.
\newblock \href {https://doi.org/10.1109/TPAMI.2019.2913372} {\path{doi:10.1109/TPAMI.2019.2913372}}.

\bibitem{Mallat2008AWay}
S.~Mallat, {A Wavelet Tour of Signal Processing: The Sparse Way}, A Wavelet Tour of Signal Processing: The Sparse Way (2008) 1--805\href {https://doi.org/10.1016/B978-0-12-374370-1.X0001-8} {\path{doi:10.1016/B978-0-12-374370-1.X0001-8}}.

\bibitem{Rhif2019WaveletReview}
M.~Rhif, A.~B. Abbes, I.~R. Farah, B.~Mart{\'{i}}nez, Y.~Sang, {Wavelet transform application for/in non-stationary time-series analysis: A review} (4 2019).
\newblock \href {https://doi.org/10.3390/app9071345} {\path{doi:10.3390/app9071345}}.

\bibitem{GARCIATREVINO2019111}
E.~S. Garc{\'{i}}a~Trevi{\~{n}}o, V.~Alarc{\'{o}}n~Aquino, J.~A. Barria, \href{http://www.sciencedirect.com/science/article/pii/S0167947318302044 https://www.sciencedirect.com/science/article/pii/S0167947318302044}{{The radial wavelet frame density estimator}}, Computational Statistics and Data Analysis 130 (2019) 111--139.
\newblock \href {https://doi.org/10.1016/j.csda.2018.08.021} {\path{doi:10.1016/j.csda.2018.08.021}}.
\newline\urlprefix\url{http://www.sciencedirect.com/science/article/pii/S0167947318302044 https://www.sciencedirect.com/science/article/pii/S0167947318302044}

\bibitem{Yu2018OnlineNetworks}
J.~J. Yu, Y.~Hou, V.~O. Li, {Online False Data Injection Attack Detection with Wavelet Transform and Deep Neural Networks}, IEEE Transactions on Industrial Informatics 14~(7) (2018) 3271--3280.
\newblock \href {https://doi.org/10.1109/TII.2018.2825243} {\path{doi:10.1109/TII.2018.2825243}}.

\bibitem{Kanarachos2017DetectingTransform}
S.~Kanarachos, S.~R.~G. Christopoulos, A.~Chroneos, M.~E. Fitzpatrick, {Detecting anomalies in time series data via a deep learning algorithm combining wavelets, neural networks and Hilbert transform}, Expert Systems with Applications 85 (2017) 292--304.
\newblock \href {https://doi.org/10.1016/j.eswa.2017.04.028} {\path{doi:10.1016/j.eswa.2017.04.028}}.

\bibitem{Bagnall2018The2018}
A.~Bagnall, H.~A. Dau, J.~Lines, M.~Flynn, J.~Large, A.~Bostrom, P.~Southam, E.~Keogh, \href{http://arxiv.org/abs/1811.00075}{{The UEA multivariate time series classification archive, 2018}} (2018).
\newline\urlprefix\url{http://arxiv.org/abs/1811.00075}

\bibitem{Shokoohi-Yekta2015OnCase}
M.~Shokoohi-Yekta, J.~Wang, E.~Keogh, {On the non-trivial generalization of Dynamic Time Warping to the multi-dimensional case}, SIAM International Conference on Data Mining 2015, SDM 2015 (2015) 289--297\href {https://doi.org/10.1137/1.9781611974010.33} {\path{doi:10.1137/1.9781611974010.33}}.

\bibitem{Smith2017CyclicalNetworks}
L.~N. Smith, {Cyclical learning rates for training neural networks}, Proceedings - 2017 IEEE Winter Conference on Applications of Computer Vision, WACV 2017 (2017) 464--472\href {https://doi.org/10.1109/WACV.2017.58} {\path{doi:10.1109/WACV.2017.58}}.

\end{thebibliography}


\end{document}